\documentclass{article}

\PassOptionsToPackage{numbers}{natbib}
\usepackage[preprint]{neurips_2024}

\usepackage[utf8]{inputenc} % allow utf-8 input
\usepackage[T1]{fontenc}    % use 8-bit T1 fonts
\usepackage{hyperref}       % hyperlinks
\usepackage{url}            % simple URL typesetting
\usepackage{booktabs}       % professional-quality tables
\usepackage{amsfonts}       % blackboard math symbols
\usepackage{nicefrac}       % compact symbols for 1/2, etc.
\usepackage{microtype}      % microtypography
\usepackage{xcolor}         % colors
\usepackage{amsmath}        % advanced math formatting
\usepackage{graphicx}       % include graphics
\usepackage{subcaption}     % subfigures and subcaptions
\usepackage{algorithm}
\usepackage{tabularx}       % for tabularx environment
\usepackage{algpseudocode}
\usepackage{float}
\title{Hyperparameter Optimisation with Practical Interpretability and Explanation Methods in Probabilistic Curriculum Learning}

\author{%
  Llewyn Salt\\
  Electrical Engineering and Computer Science\\
  University of Queensland\\
  Brisbane, Australia\\
  \texttt{llewyn.salt@gmail.com} \\
  \And
  Marcus Gallagher\\
  Electrical Engineering and Computer Science\\
  University of Queensland\\
  Brisbane, Australia\\
  \texttt{marcusg@eecs.uq.edu.au} \\
}

\begin{document}

\maketitle

\begin{abstract}
Hyperparameter optimisation (HPO) is crucial for achieving strong performance in reinforcement learning (RL), as RL algorithms are inherently sensitive to hyperparameter settings. Probabilistic Curriculum Learning (PCL) is a curriculum learning strategy designed to improve RL performance by structuring the agent's learning process, yet effective hyperparameter tuning remains challenging and computationally demanding. In this paper, we provide an empirical analysis of hyperparameter interactions and their effects on the performance of a PCL algorithm within standard RL tasks, including point-maze navigation and DC motor control. Using the AlgOS framework integrated with Optuna’s Tree-Structured Parzen Estimator (TPE), we present strategies to refine hyperparameter search spaces, enhancing optimisation efficiency. Additionally, we introduce a novel SHAP-based interpretability approach tailored specifically for analysing hyperparameter impacts, offering clear insights into how individual hyperparameters and their interactions influence RL performance. Our work contributes practical guidelines and interpretability tools that significantly improve the effectiveness and computational feasibility of hyperparameter optimisation in reinforcement learning.
\end{abstract}

\section{Introduction}

Hyperparameter optimisation (HPO) is a crucial component of the machine learning pipeline due to the significant impact hyperparameters have on model performance. Machine learning models possess numerous configurable settings, called hyperparameters, such as learning rates, hidden layer sizes, optimisation algorithms and their settings, regularisation parameters, and more. Unlike model parameters, hyperparameters are not directly learned during training but rather set prior to optimisation. These hyperparameters are often codependent, numerous, and complex to tune~\cite{salt2019parameter}. Thus, effective hyperparameter optimisation significantly enhances model performance and generalisation~\cite{JMLR:v13:bergstra12a, zhang2021importance, paine2020hyperparameter, hutter2015beyond, frazier2018tutorial}.

The optimisation of hyperparameters typically involves navigating high-dimensional, multimodal spaces that are prone to local minima. Common methods for hyperparameter optimisation include random search, grid search~\cite{bergstra2012random}, Bayesian optimisation~\cite{snoek2012practical}, and evolutionary algorithms~\cite{real2017large}. However, these approaches often require considerable computational resources and extensive evaluation budgets. Additionally, hyperparameter spaces frequently comprise mixtures of discrete and continuous variables, prohibiting gradient-based optimisation.

Reinforcement learning (RL), a subfield focused on training agents to make decisions by maximising cumulative rewards, poses particular challenges for hyperparameter optimisation due to the inherent complexity and sensitivity of RL algorithms to hyperparameters such as discount factors, exploration rates, and neural network architectures~\cite{li2022hyperband}. Successful RL algorithms, including AlphaGo~\cite{silver2016mastering} and AlphaStar~\cite{vinyals2019grandmaster}, illustrate the crucial role of HPO, yet these successes often rely heavily on vast computational resources unavailable to many practitioners. Furthermore, existing literature frequently discusses HPO superficially, lacking detailed methodological insights into interpreting hyperparameter interactions and effects, often relying on automated frameworks like AutoML~\cite{he2021automl} without deeper exploration.

In this paper, we focus on hyperparameter optimisation and analysis within probabilistic curriculum learning (PCL)~\cite{salt2025probabilisticcurriculumlearninggoalbased}, as it is a reinforcement learning algorithm which are known for their sensitivity to hyperparameters~\cite{henderson2018deep, islam2017reproducibility}. We present empirical results conducted on RL benchmark tasks, such as point-maze navigation and DC motor control, using the AlgOS framework~\cite{salt2025}. AlgOS provides valuable HPO support, including unified interfaces for hyperparameter bounding, structured logging of experimental metadata for post-hoc analysis, and direct integration with the Optuna hyperparameter optimisation framework~\cite{optuna_2019}.

Our primary contributions include:

\begin{enumerate} 
  \item A thorough empirical analysis of hyperparameter impacts and interactions within the PCL algorithm, including demonstrating empirically hyperparameters have the described effects. 
  \item Practical guidelines and strategies to refine hyperparameter bounds, improving both optimisation efficiency and model performance. 
  \item Development and demonstration of a novel SHAP-based interpretability approach specifically designed for analysing hyperparameter importance in reinforcement learning tasks. 
\end{enumerate}

Through these contributions, we aim to provide clear, actionable insights into hyperparameter optimisation and interpretation, addressing a crucial gap in current reinforcement learning and hyperparameter optimisation research.

\section{Background}

\subsection{Probabilistic Curriculum Learning}
Probabilistic Curriculum Learning (PCL) is a curriculum learning strategy that structures the learning process of reinforcement learning agents by introducing tasks of increasing complexity. PCL employs a probabilistic approach to sample tasks from a distribution, allowing the agent to learn from simpler tasks before progressing to more complex ones. This method has been shown to improve the efficiency and effectiveness of the learning process in continuous environments~\cite{salt2025probabilisticcurriculumlearninggoalbased}.
PCL uses a deep mixture density network~\cite{bishop1994mixture, makansi2019overcoming, girbau2021multiple} (MDN) to model the likelihood of reaching a goal state given the current state and action. The MDN is trained to predict the distribution of possible next states, allowing the agent to sample from this distribution to select actions that are more likely to lead to successful outcomes. This probabilistic approach enables the agent to adaptively adjust its learning strategy based on the complexity of the tasks it encounters.

\subsection{Hyperparameter Optimisation via Tree-Structured Parzen Estimators}

Bayesian optimisation techniques, such as the Tree-Structured Parzen Estimator (TPE), have become popular for hyperparameter optimisation due to their efficiency and ability to handle complex hyperparameter spaces~\cite{bergstra2011algorithms}. TPE iteratively builds probabilistic models of the objective function based on previously observed hyperparameter configurations, efficiently guiding the search toward promising regions of the parameter space.

The TPE algorithm starts by sampling initial hyperparameter configurations, often via random or Latin hypercube sampling, to evaluate the objective function. It then constructs two probabilistic models: one density function l(x)l(x), estimating configurations with low observed loss, and another density function g(x)g(x), estimating configurations associated with high observed loss. The algorithm uses the Expected Improvement (EI) criterion defined as:

\begin{equation} 
  \text{EI}(x) = \frac{l(x)}{g(x)} 
\end{equation}

This criterion preferentially selects hyperparameter configurations predicted to yield better (lower loss) results. At each iteration, the TPE algorithm maximises the EI to determine the next hyperparameter configuration to evaluate. The process continues until a predefined stopping criterion is met.

TPE's primary advantages include scalability, flexibility, and its non-parametric nature. However, TPE can become computationally intensive for very high-dimensional parameter spaces, and performance may be sensitive to the choice of initial sampling methods and hyperparameter space definitions~\cite{bergstra2011algorithms}. In this work, we specifically utilise TPE as implemented by Optuna due to its robustness, scalability, and effective integration within the AlgOS experimental framework.

\subsection{Interpretability with SHAP (SHapley Additive exPlanations)}

SHapley Additive exPlanations (SHAP) is a popular interpretability method that uses Shapley values from cooperative game theory to allocate contributions to model predictions among individual input features~\cite{NIPS2017_7062}. Shapley values ensure a fair distribution of importance among features by averaging marginal contributions across all possible feature coalitions. Unlike traditional sensitivity analyses, which often fail to capture nonlinear interactions or complex dependencies among variables, SHAP provides a rigorous theoretical basis for interpreting the contributions of each feature. SHAP has been effectively employed in diverse fields including healthcare diagnostics, environmental modelling, concrete strength estimation, diesel yield optimisation, and polymer characteristic predictions~\cite{nohara2022explanation,aldrees2024evaluation, ekanayake2022novel, qi2025study, agrawal2024prediction}.

In the context of hyperparameter optimisation, SHAP offers a powerful way to elucidate the relative importance and interactions among hyperparameters on model performance, something often neglected or superficially addressed in reinforcement learning literature. Our work presents a novel SHAP-based analysis methodology explicitly tailored to evaluating hyperparameter contributions in reinforcement learning, demonstrating insights that significantly enhance understanding and improve optimisation procedures.

\section{Methodology}
\label{ssec:goal:methodology}
All experiments were conducted using the AlgOS framework~\cite{salt2025} to evaluate the Probabilistic Curriculum Learning (PCL) algorithm~\cite{salt2025probabilisticcurriculumlearninggoalbased}. We utilised the Soft Actor-Critic (SAC) algorithm from Stable Baselines 3~\cite{stable-baselines3} as the reinforcement learning agent, testing PCL’s effectiveness in two continuous control tasks: a DC Motor control environment, a simple single-input-single-output (SISO) problem without obstacles, and a more complex point-robot navigation task~\cite{gymnasium_robotics2023github}, presenting a multi-input-multi-output (MIMO) control challenge involving obstacles.

AlgOS provides an optimisation interface via Optuna~\cite{optuna_2019}, which allows us to tune the numerous hyperparameters of both PCL and SAC using a tree parzen estimator~\cite{watanabe2023tree}, a form of Bayesian optimisation requiring fewer samples~\cite{salt2019parameter}. Due to compute resourcing constraints, we optimise over 150000 steps for all environments which is a relatively small number when compared to the number of steps used in the literature~\cite{raffin2021smooth}.

Hyperparameter spaces included both continuous and discrete parameters, clearly bounded and categorised via AlgOS. Tables~\ref{tab:MDNModelparameters},\ref{tab:MDNModel2-parameters}, and\ref{tab:SACWeightedMDNExperiment} summarise the hyperparameter bounds used across initial, intermediate, and final optimisation phases.

We define our objective function as the coverage of goals the agent was able to achieve in the environment. Coverage is defined as the number of times an agent was successful in its last evaluation divided by the total number of evaluations $f_{objective} = \frac{1}{N}\sum_N{g_{success}}$. Evaluations were conducted every 30,000 steps, with each goal assessed four times. Goals for the point-maze navigation task were predefined, whereas goals for the DC motor task were linearly spaced between -1 and 1 at intervals of 0.1.

\section{Distribution and Surface Plot Analysis}

To systematically refine hyperparameter bounds, we employed histograms to visualise distributions of successful hyperparameter values and identify skewness or uniformity, suggesting expansions or contractions of search spaces. Additionally, we used two-dimensional surface plots to visualise pairwise hyperparameter interactions.

\begin{figure}[htbp]
  \centering
  \begin{subfigure}[b]{0.45\textwidth}
      \includegraphics[width=\textwidth, height=4cm]{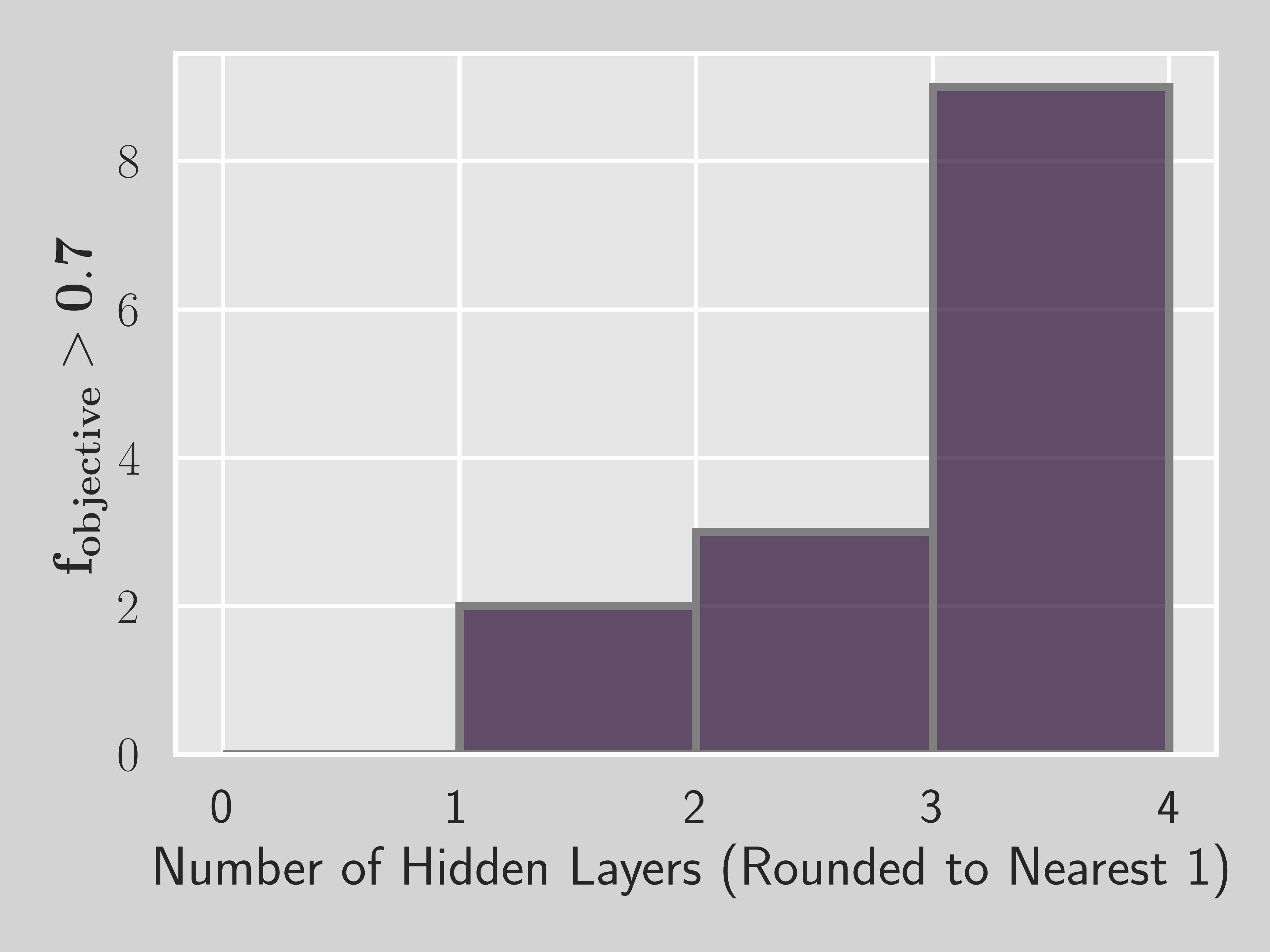}
      \caption{Number of Hidden Layers (MDN)}
      \label{fig:MDNInferenceModel-hidden-layers-length}
  \end{subfigure}
  \hfill
  \begin{subfigure}[b]{0.45\textwidth}
    \includegraphics[width=\textwidth, height=4cm]{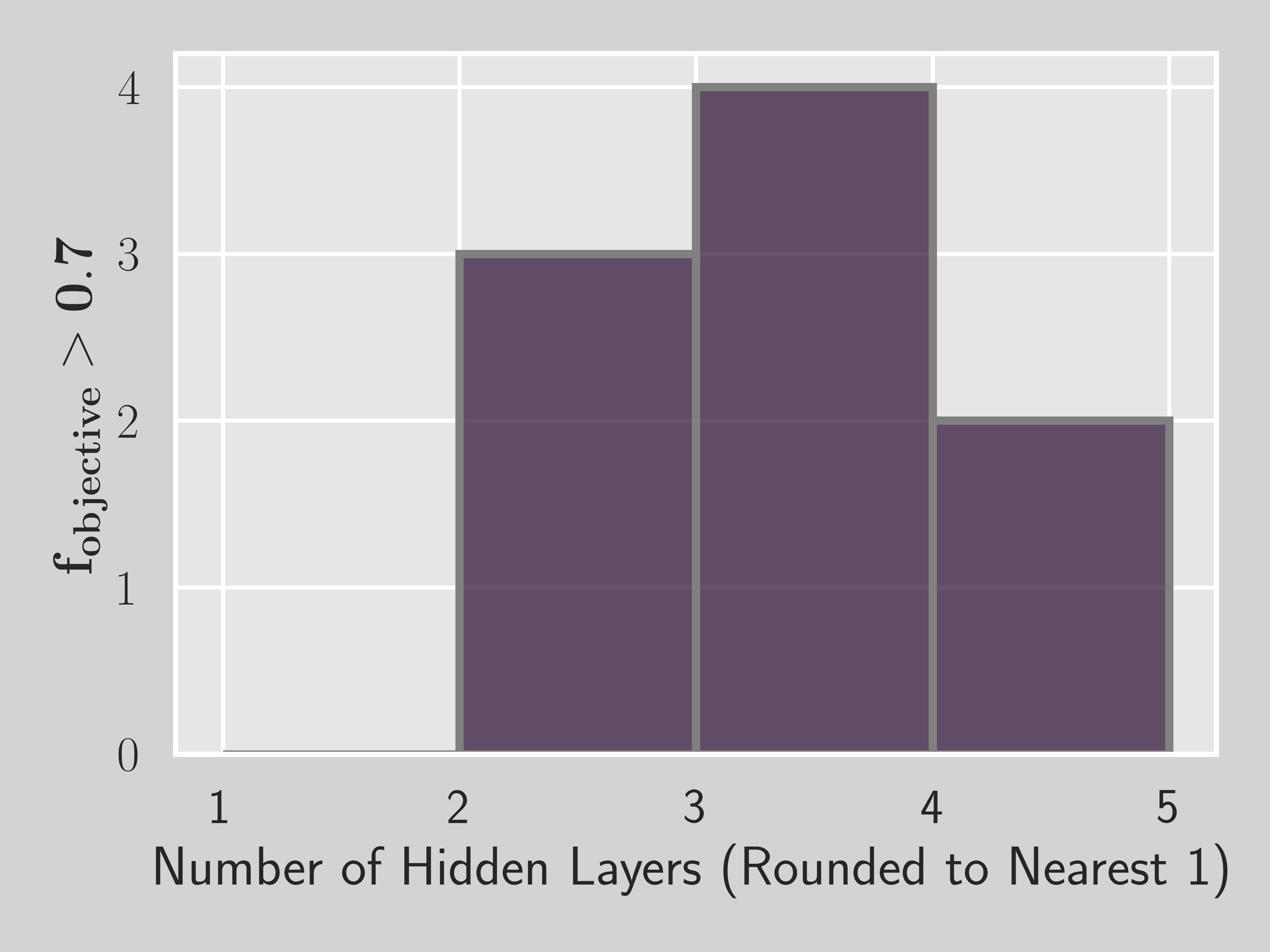}
    \caption{Number of Hidden Layers (SAC)}
    \label{fig:SACSB3Agent-net-arch-length}
  \end{subfigure}
  
  \begin{subfigure}[b]{0.45\textwidth}
    \includegraphics[width=\textwidth, height=4cm]{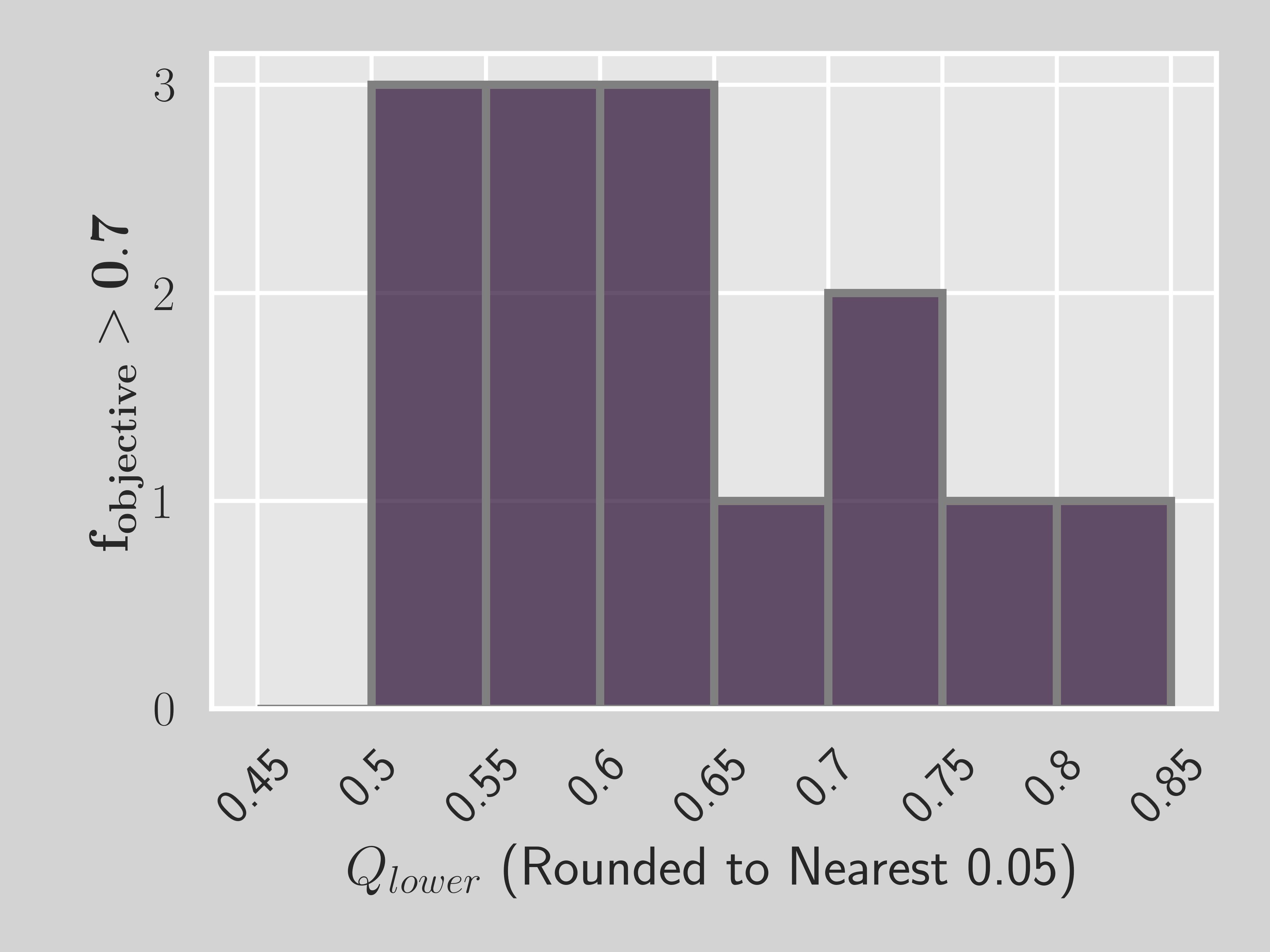}
    \caption{$Q_{lower}$ (MDN)}
    \label{fig:PDESelector-q-lower}
  \end{subfigure}
  \hfill
  \begin{subfigure}[b]{0.45\textwidth}
    \includegraphics[width=\textwidth, height=4cm]{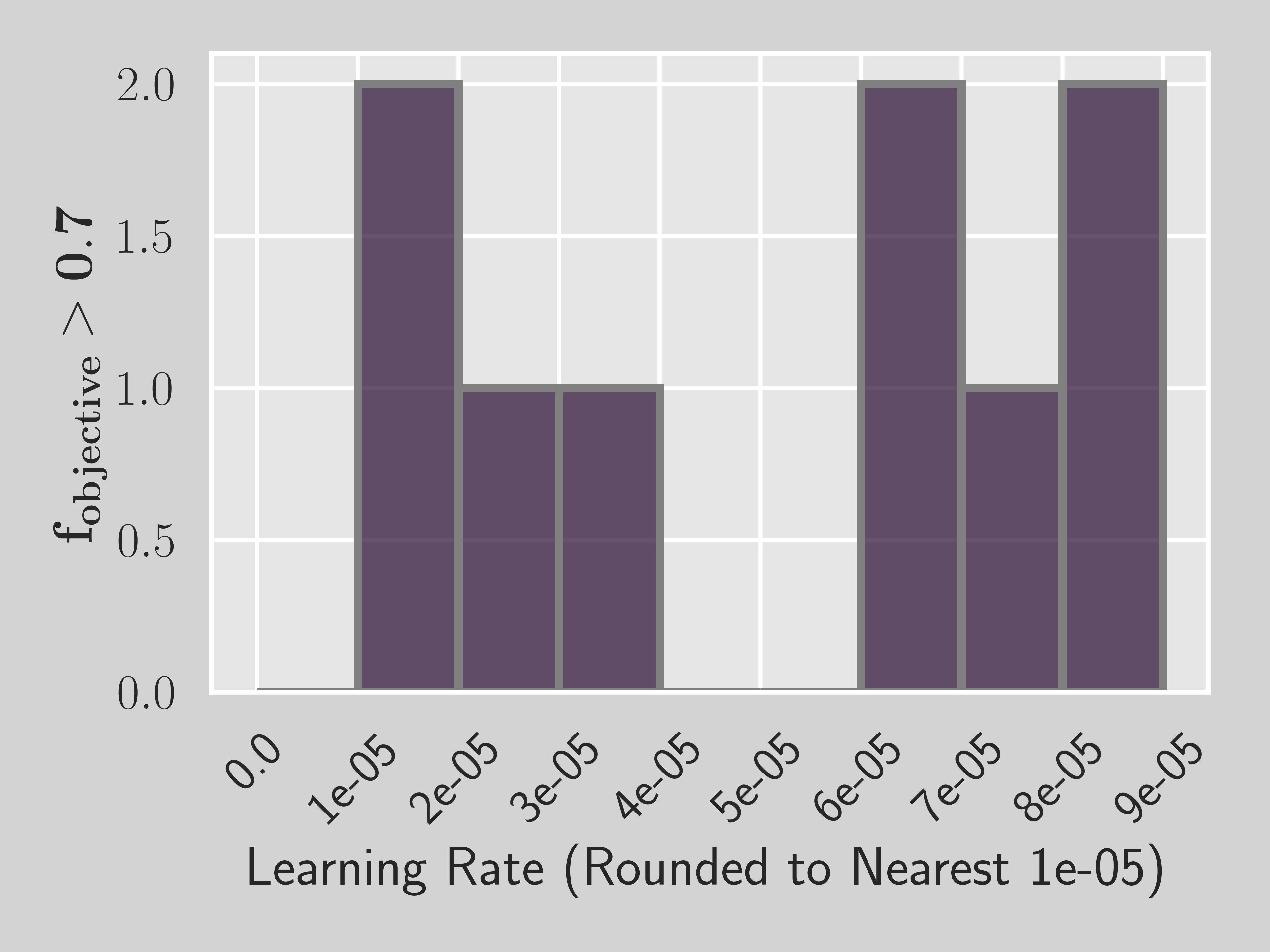}
    \caption{Learning Rate (SAC)}
    \label{fig:lr:sac}
  \end{subfigure}
  \caption{Histograms for hyperparameters of the PCL model where $f_{objective}>0.7$.}
  \label{fig:initial-histograms}
\end{figure}

Figure~\ref{fig:initial-histograms} shows some hyperparameters distributions for experiments conducted with the hyperparameter bounds in Table~\ref{tab:MDNModelparameters}. Figure~\ref{fig:MDNInferenceModel-hidden-layers-length} shows that the number of hidden layers for the MDN is negatively skewed, indicating that the upper and lower bounds should be increased. The SAC agent seems to perform best with two or tree hidden layers as shown in Figure~\ref{fig:SACSB3Agent-net-arch-length}. Figure~\ref{fig:PDESelector-q-lower} shows that the lower quantile is positively skewed indicating that the upper and lower bounds should be decreased. Figure~\ref{fig:lr:sac} shows that the learning rate for the SAC agent is relatively uniform across the bounds indicating that they should be expanded. 

\begin{figure}[htbp]
  \centering
  \begin{subfigure}[b]{0.45\textwidth}
      \includegraphics[width=\textwidth, height=4cm]{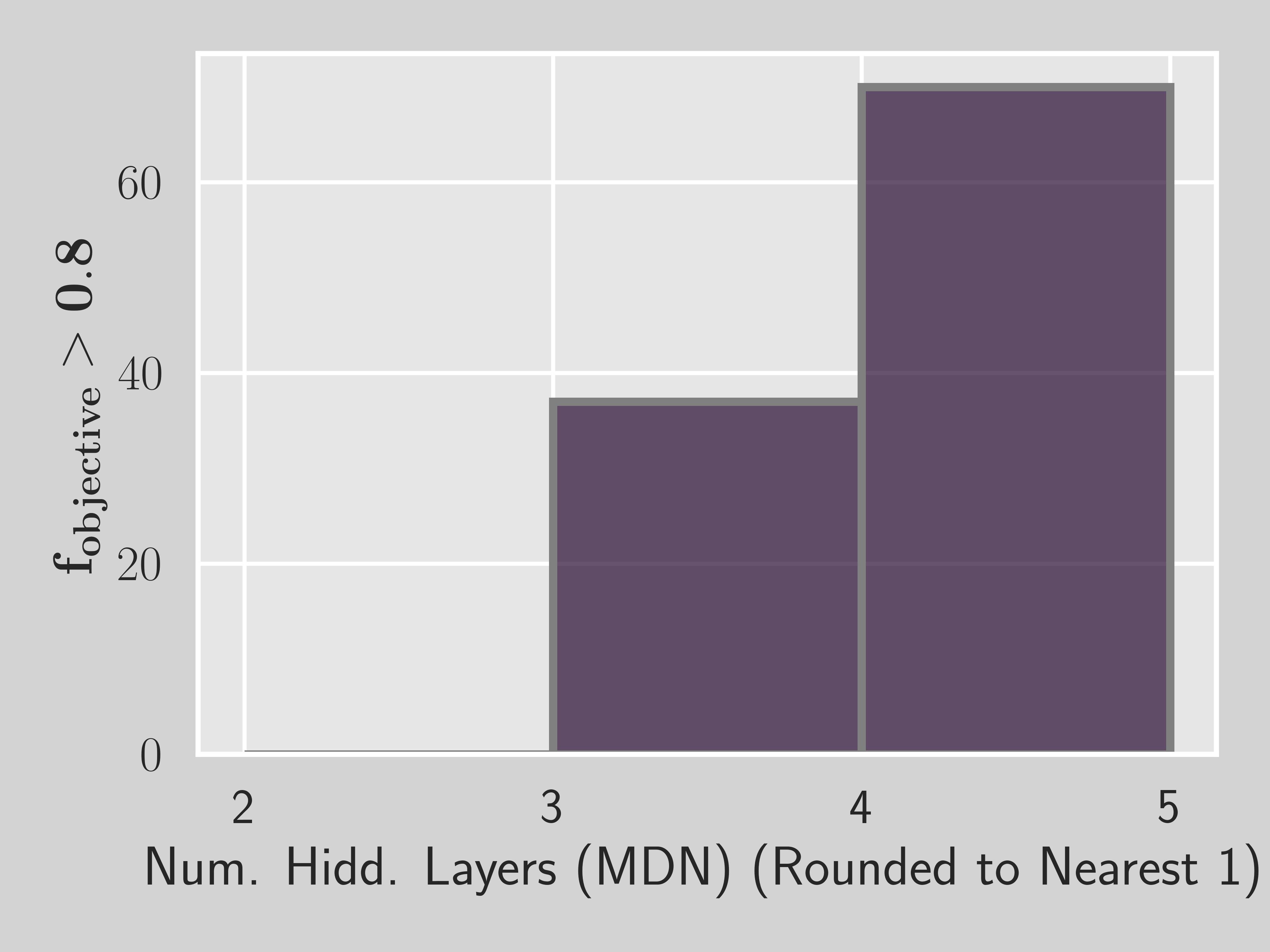}
      \caption{Number of Hidden Layers (MDN)}
      \label{fig:MDNInferenceModel-hidden-layers-length:intermediate}
  \end{subfigure}
  \hfill
  \begin{subfigure}[b]{0.45\textwidth}
    \includegraphics[width=\textwidth, height=4cm]{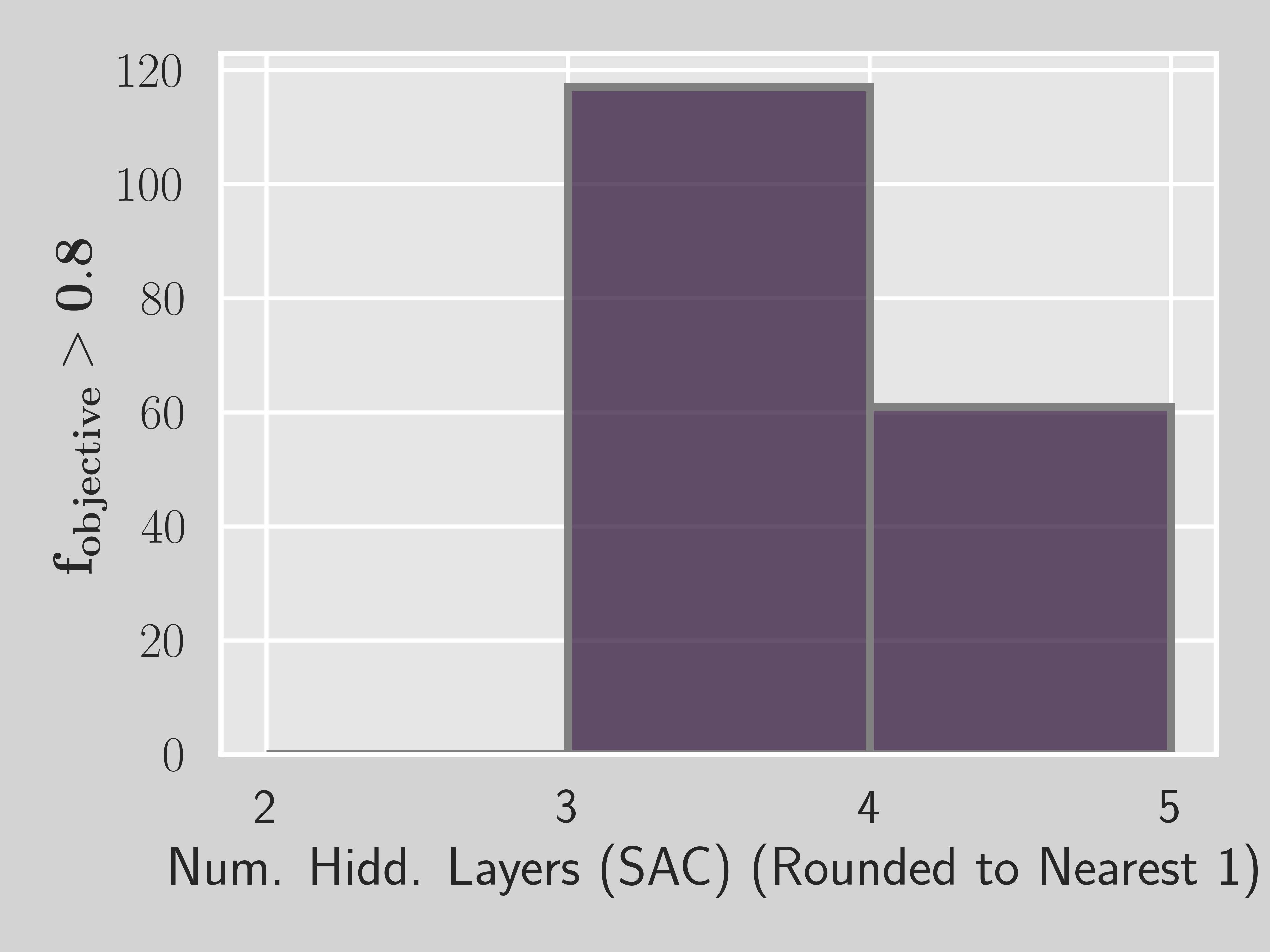}
    \caption{Number of Hidden Layers (SAC)}
    \label{fig:SACSB3Agent-net-arch-length:intermediate}
  \end{subfigure}
  
  \begin{subfigure}[b]{0.45\textwidth}
    \includegraphics[width=\textwidth, height=4cm]{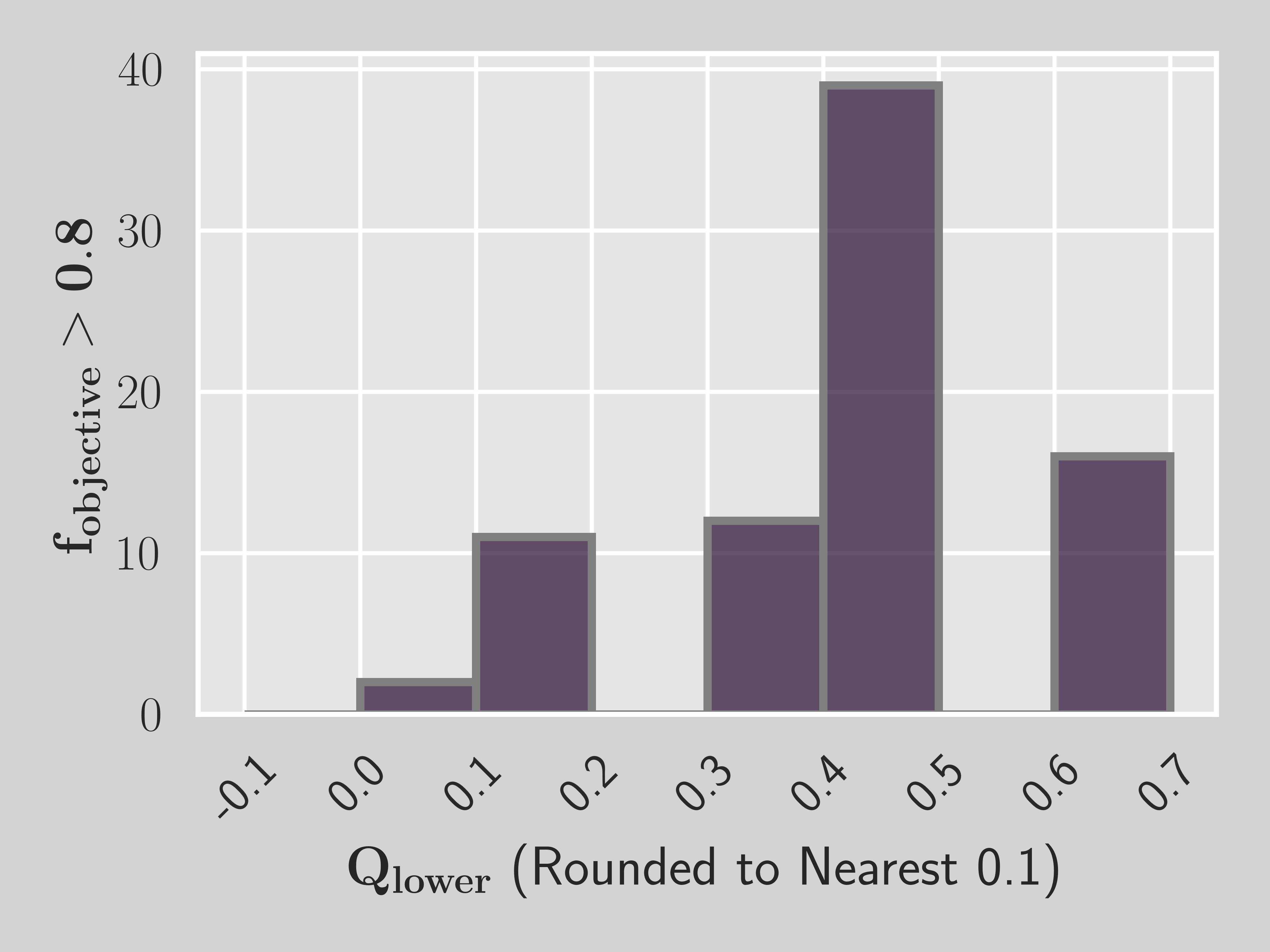}
    \caption{$Q_{lower}$ (MDN)}
    \label{fig:PDESelector-q-lower:intermediate}
  \end{subfigure}
  \hfill
  \begin{subfigure}[b]{0.45\textwidth}
    \includegraphics[width=\textwidth, height=4cm]{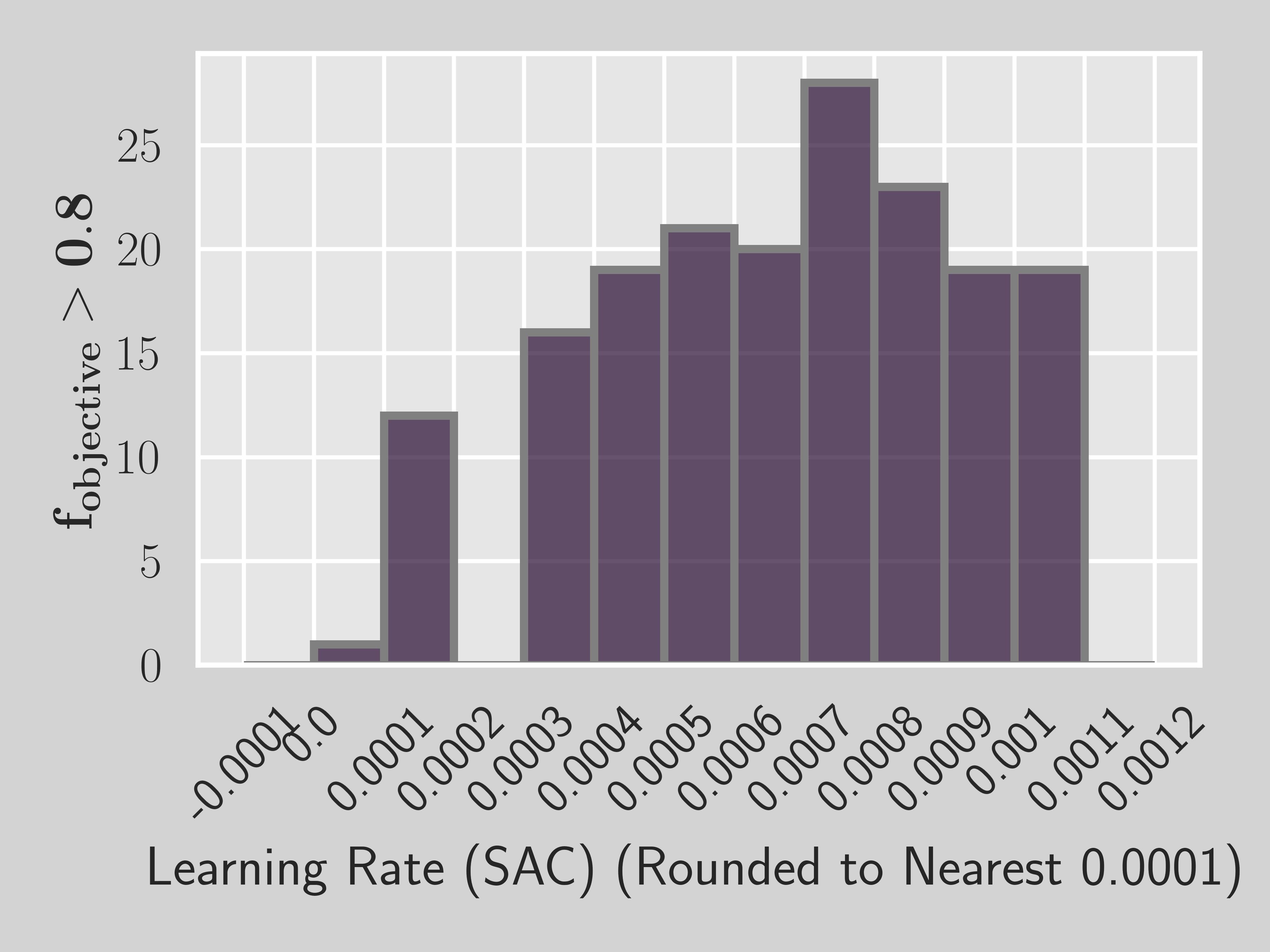}
    \caption{Learning Rate (SAC)}
    \label{fig:lr:sac:intermediate}
  \end{subfigure}
  \caption{Histograms for hyperparameters of the PCL model where $f_{objective}>0.8$.}
  \label{fig:initial-histograms:intermediate}
\end{figure}

After adjusting based on our initial analysis, we can investigate the hyperparameter distributions again. Figure~\ref{fig:MDNInferenceModel-hidden-layers-length:intermediate} shows that the number of hidden layers for the MDN is still negatively skewed, however as the bounds are 3 to 4 and 3 still performs well expanding the bounds could be beneficial. Figure~\ref{fig:SACSB3Agent-net-arch-length:intermediate} shows that the SAC agent is positively skewed, but the same rationale as the MDN can be applied, and the bounds can be expanded. Figure~\ref{fig:PDESelector-q-lower:intermediate} shows that the lower quantile performs overwhelmingly the best between 0.4 and 0.5, indicating that shifting the lower bound lower was the correct move. Figure~\ref{fig:lr:sac:intermediate} shows that the learning rate for the SAC agent is relatively uniform across the bounds indicating that they should be expanded.

The important thing to remember with histograms is that they ignore the co-dependence of hyperparameters and that the interplay of them could lead us to constrain the space to a local optima. However, depending on computational resources, this can be attractive as it can lead to interesting results with fewer samples. 

\begin{figure}[htbp]
  \begin{center}
    \begin{subfigure}[b]{0.45\textwidth}
      \includegraphics[width=\textwidth, height=4cm]{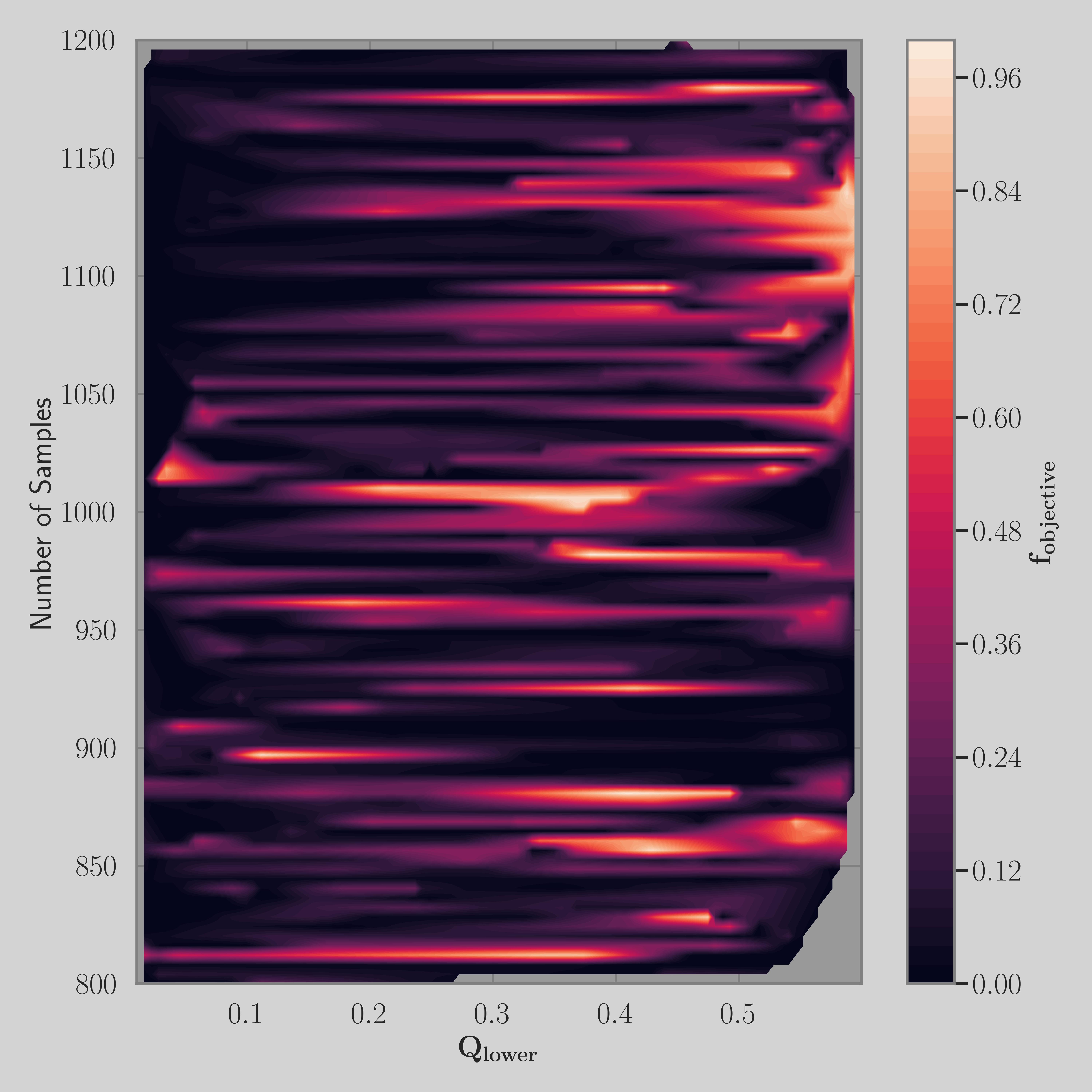}
      \caption{$Q_{lower}$ vs Number of Samples.}
      \label{fig:surface:q-lower-samples}
    \end{subfigure}
    \hfill
    \begin{subfigure}[b]{0.45\textwidth}
      \includegraphics[width=\textwidth, height=4cm]{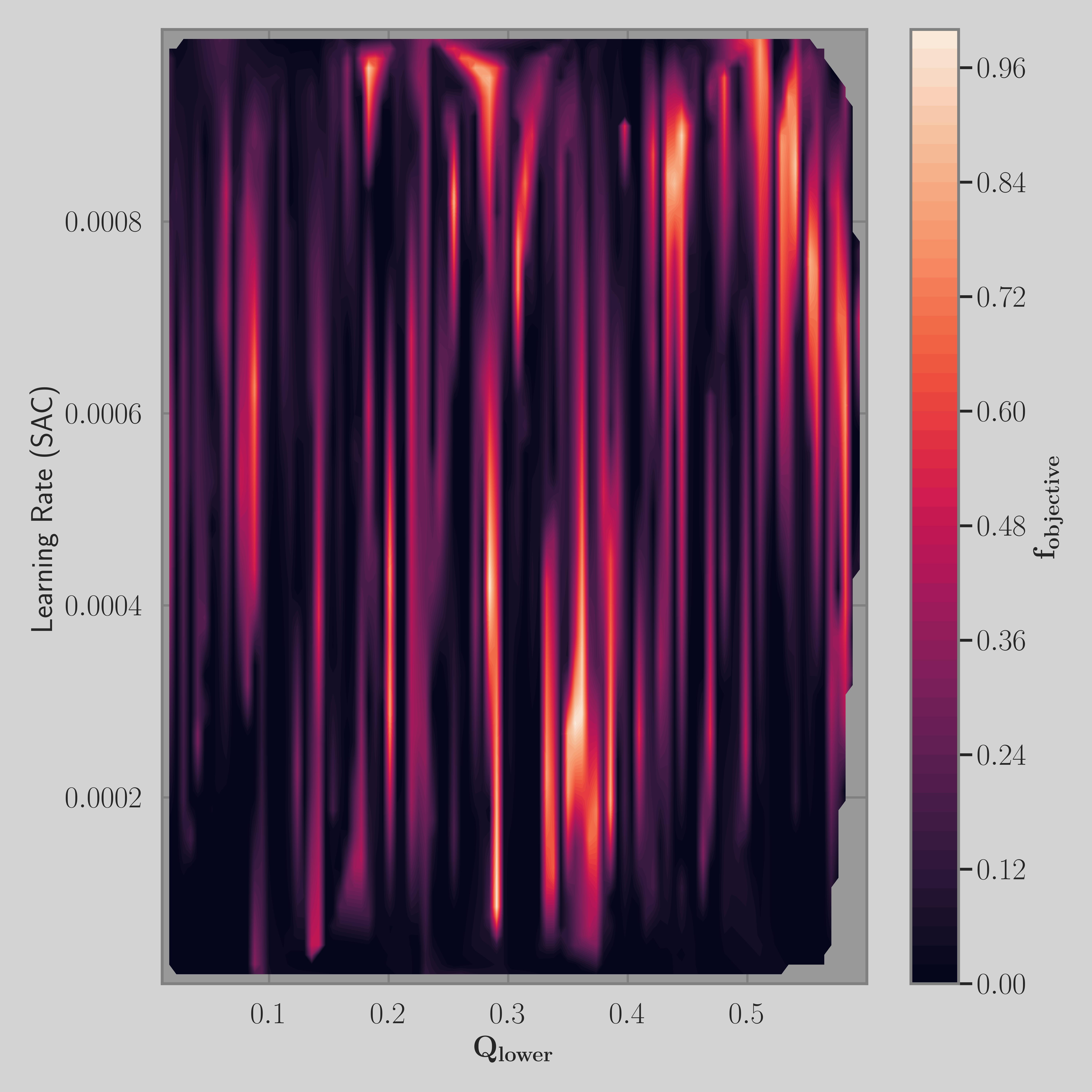}
      \caption{$Q_{lower}$ vs Learning Rate (SAC)}
      \label{fig:surface:qlower-lr}
    \end{subfigure}
  
    \begin{subfigure}[b]{0.45\textwidth}
        \includegraphics[width=\textwidth, height=4cm]{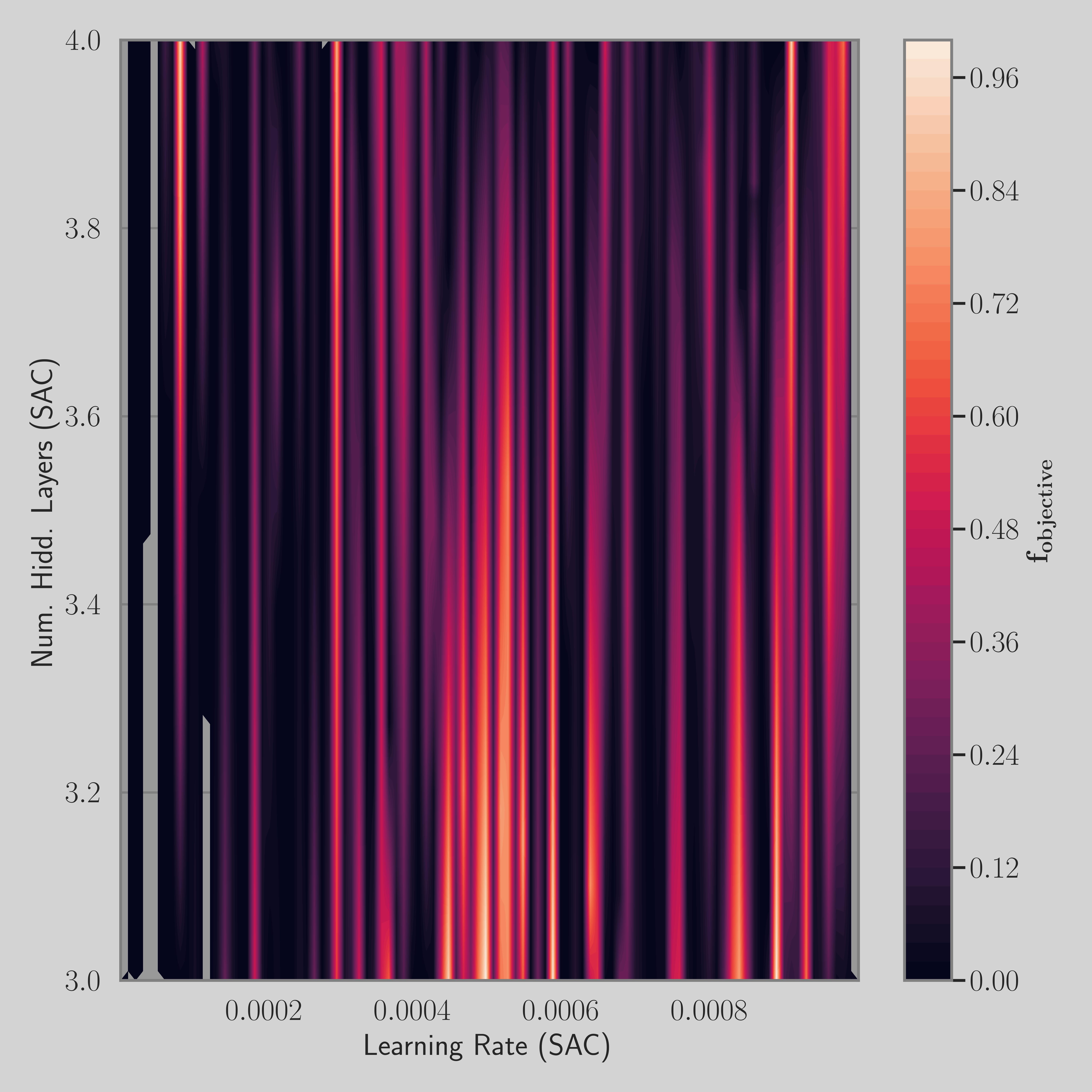}
        \caption{Learn. Rate (SAC) vs Hidden Layers (SAC)}
        \label{fig:surface:lr-hl}
    \end{subfigure}
    \hfill
    \begin{subfigure}[b]{0.45\textwidth}
      \includegraphics[width=\textwidth, height=4cm]{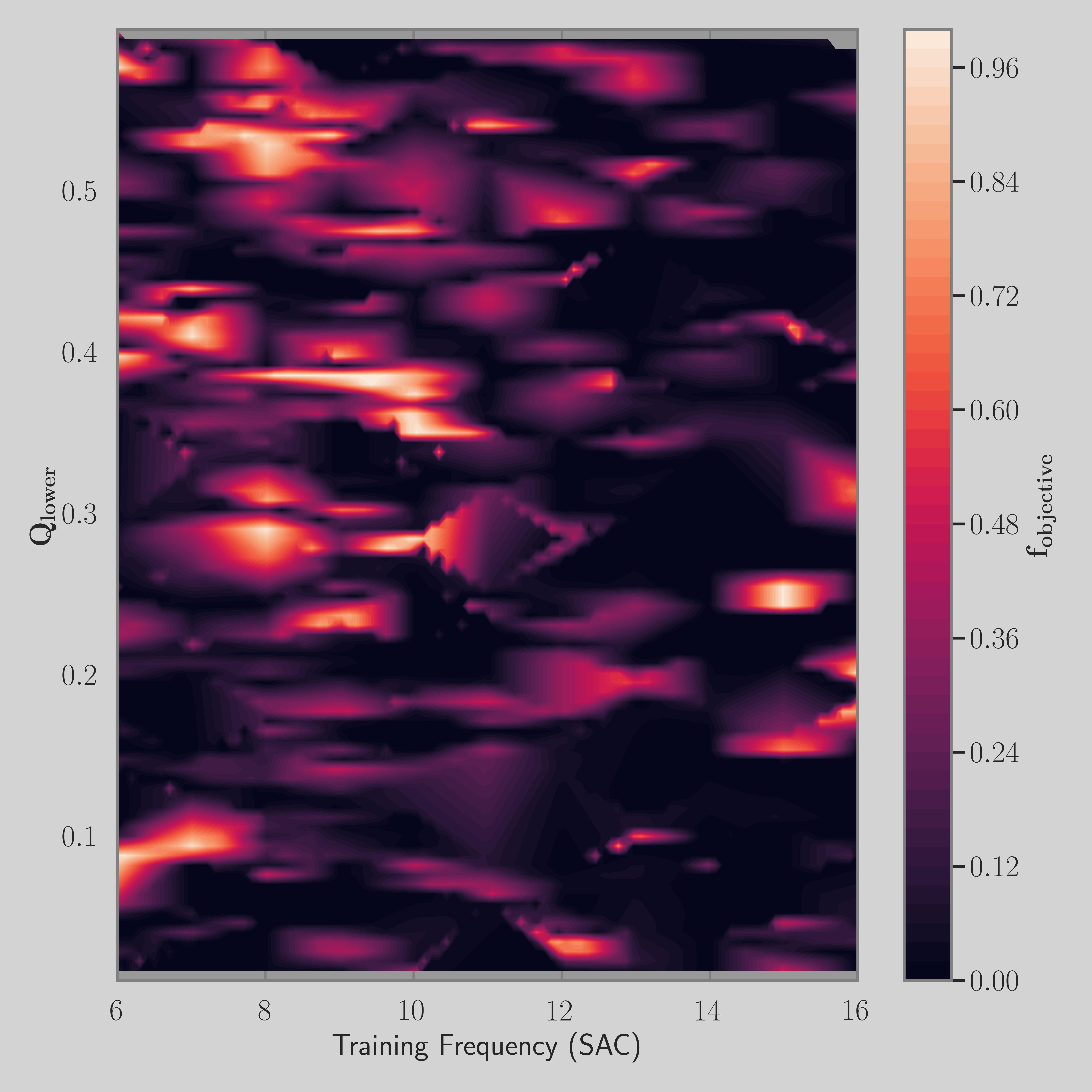}
      \caption{Training Freq. (SAC) vs $Q_{lower}$}
      \label{fig:surface:tfs-qlower}
    \end{subfigure}
    \caption{2D surfaces of Most Correlated when $f_{objective}>0.8$ PCL parameters.}
    \label{fig:surfaces:lr}
  \end{center}
\end{figure}

We can use a cross correlation matrix on the inputs using the Pearson coefficient to see if there are any correlation between the hyperparameters. Figure~\ref{goal:shap:correlation_matrix} shows that the correlations between inputs are relatively low when there is no filtering on the objective value. However, the strongest positive correlation between $Q_{lower}$ and number of samples. The strongest negative correlation between the number of hidden layers in SAC and SAC's learning rate, $\lambda_1$ and number of mixtures, and training frequency of SAC and the learning rate of the MDN. The plot isn't particularly informative given the low values, but it does show that the hyperparameters are not strongly linearly correlated. However, we can use the correlated hyperparameters in our SHAP codependency analysis.
\begin{figure}
  \centering
      \includegraphics[width=0.65\textwidth, scale=0.65]{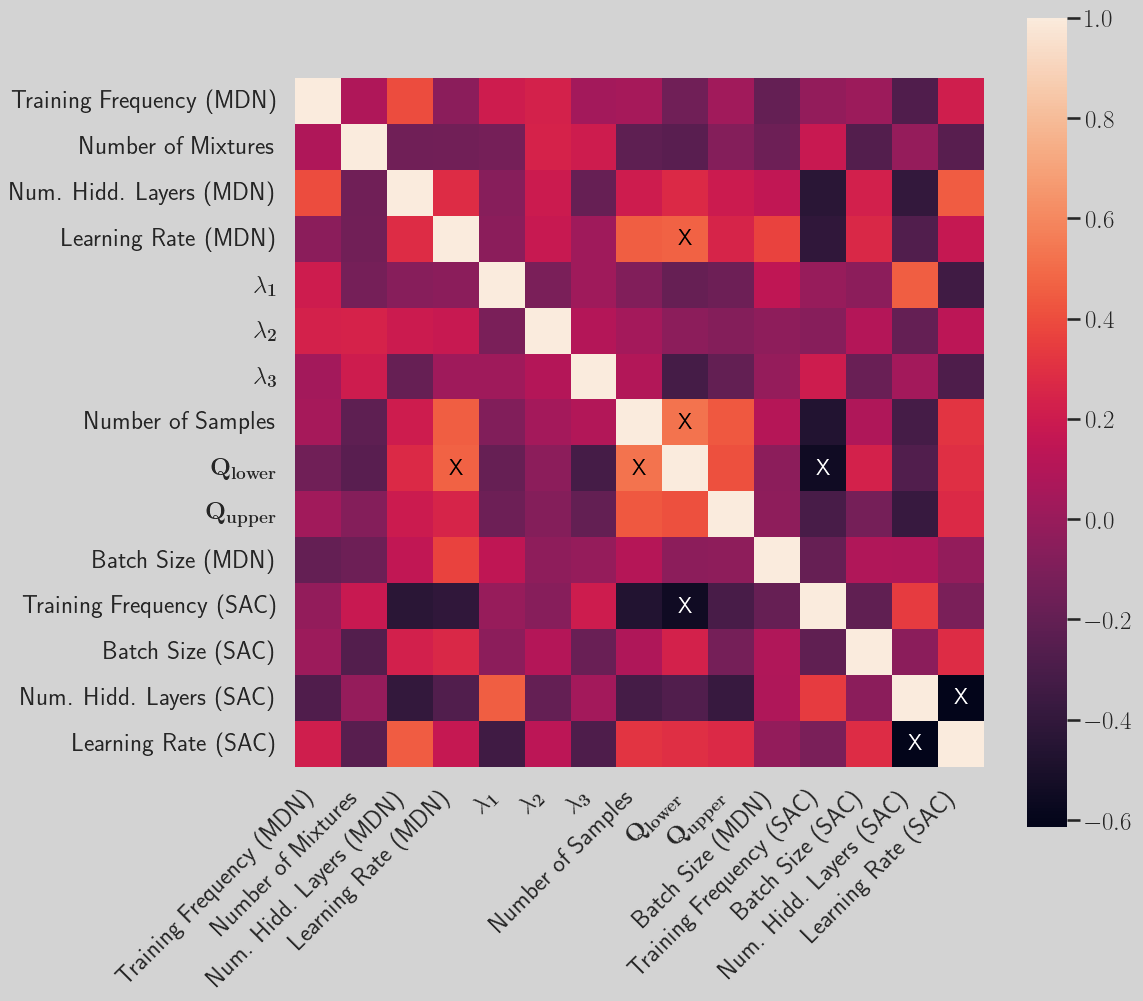}
      \caption{Correlation matrix of hyperparameters for the PCL in the PointMaze environments where $f_{objective}>0.8$.}
      \label{goal:shap:correlation_matrix_good}
  \end{figure}

Figure~\ref{goal:shap:correlation_matrix_good} shows the correlation matrix when we filter out results that do now meet the condition $f_{objective}>0.8$. We can see strong correlations here, the black Xs show the two most positively correlated hyperparameter sets, $Q_{lower}$ and Learning Rate (MDN), and $Q_{lower}$ and Number of Samples. The white Xs show the two strongest negative correlations which are Training Frequency (SAC) and $Q_{lower}$, and Learning Rate (SAC) and Number of Hidden Layers (SAC). We can use these values to investigate the hyperparameters by plotting them as a two-dimensional surface plot.

\begin{figure}[htbp]
  \centering
  \includegraphics[width=0.75\textwidth, height=4.5cm]{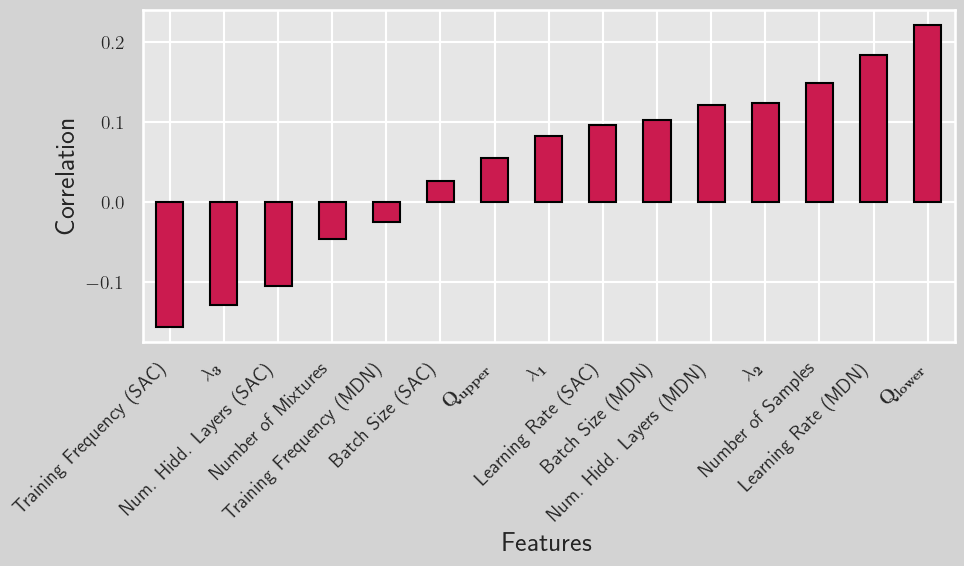}
  \caption{Correlation of hyperparameters and the objective value for the PCL in the PointMaze environments.}
  \label{goal:shap:correlation_matrix_y}
\end{figure}

Figures~\ref{fig:surface:q-lower-samples},~\ref{fig:surface:qlower-lr}, and ~\ref{fig:surface:tfs-qlower} show relatively striated surfaces, indicating that fixed values for a given parameter perform well for across the range of the other. All plots in Figure~\ref{fig:surfaces:lr} show that the optimisation space is not particularly smooth. Figure~\ref{fig:surface:tfs-qlower} shows that the larger $Q_{lower}$ and lower training frequency of SAC performs better.

Another approach would be to observe the correlation between the hyperparameters and the objective value. Figure~\ref{goal:shap:correlation_matrix_y} shows the correlation between the hyperparameters and the objective value. The strongest positive correlation is between $Q_{lower}$ and the objective value, and the strongest negative correlation is between the training frequency of SAC and the objective value. This indicates that filtering out the easy goals is important for the performance of the agent and that updating the agent less frequently causes the agent to perform worse. Interestingly, the more hyperparameters that are more positively correlated with the objective value are associated with the MDN or curriculum component of PCL, highlighting the importance of the curriculum in the performance of the agent. 

\section{SHAP Analysis}
To rigorously explain and interpret hyperparameter impacts, we developed a novel approach using SHapley Additive exPlanations (SHAP). Hyperparameter configurations and corresponding objective values from Optuna trials were aggregated and modelled using a Scikit-Learn Random Forest regressor, chosen specifically for its ability to capture nonlinear interactions effectively and robustly even with limited data samples (approximately 800). Data was split into 80:20 train:test subsets, and performance was validated through mean squared error (MSE) on the test set. 

To maximise the amount of data points available we can aggregate experiments that share the same hyperparameters. This will hopefully minimise some of the variance due to the relatively low number of samples (in the hundreds rather than thousands). The SHAP explanation is useful because it tells us if a hyperparameter is having a positive or negative effect on the objective value as well as the magnitude of the hyperparameter's value (low or high relative to its bounds). We can also then investigate how hyperparameters interact with each other through SHAP dependence plots which show the hyperparameter impact, the hyperparameter value, and colourises the data points based on the value of the interaction hyperparameter.

\begin{figure}[htbp]
  \begin{center}
      \includegraphics[width=0.55\textwidth, scale=0.55]{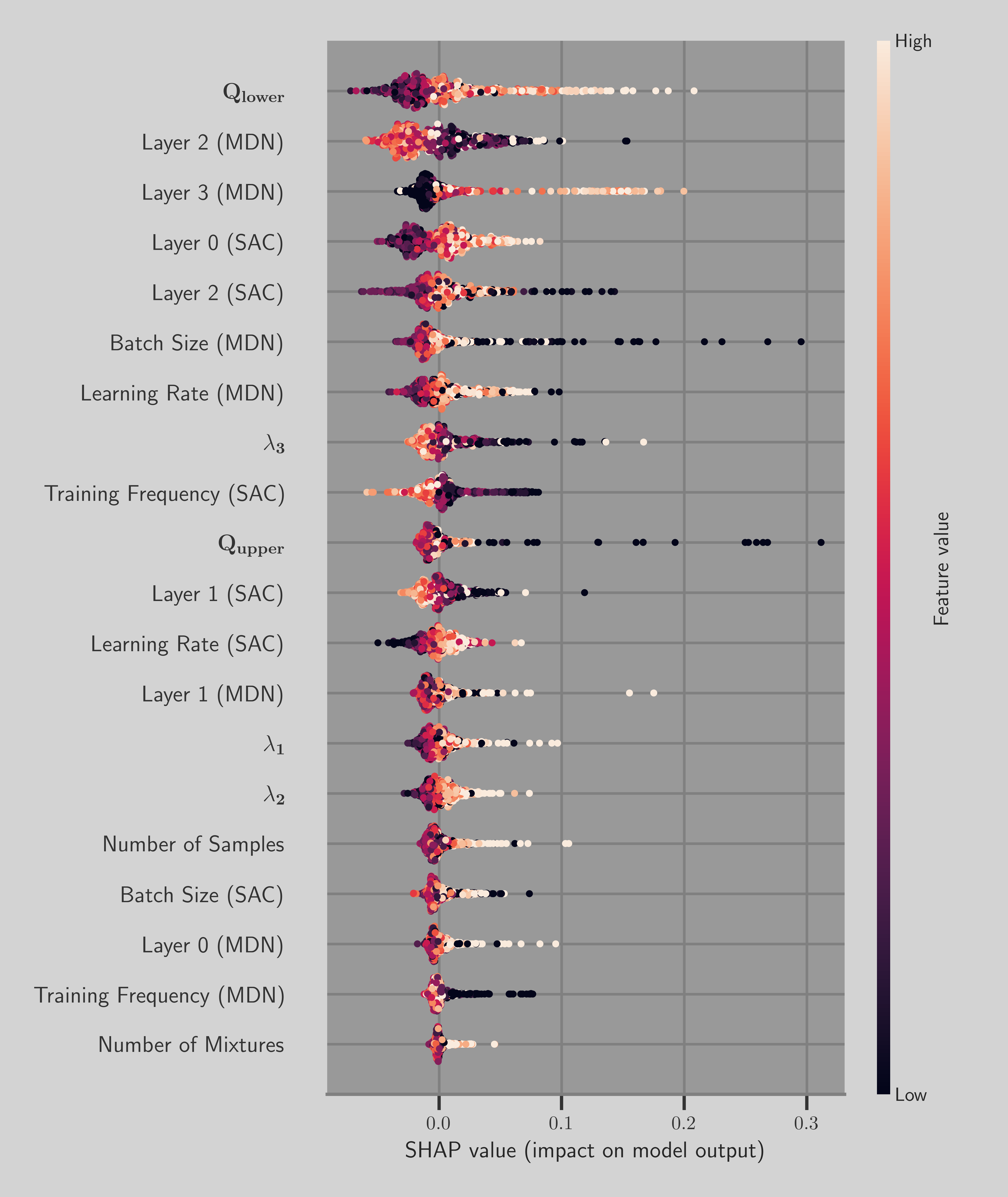}
      \caption{Combined SHAP summary plot for the MDN Curriculum with a SAC agent in the PointMaze environments.}
      \label{goal:shap:combined_summary}
  \end{center}
\end{figure}

Figure~\ref{goal:shap:combined_summary} shows the combined SHAP summary plot for the PCL in the PointMaze environment using multiple maze configurations. From the plot we can see the utility of the SHAP explainer for hyperparameter tuning. The hyperparameters are ordered by importance. It provides a SHAP value and indicates whether the hyperparameter is having a positive or negative effect on the objective value. It also shows the magnitude of the value of the hyperparameter. This is useful if the same colours sit to the left or the right of the 0 axis. If a lot of high or low values are sitting in the positive space then perhaps the hyperparameter bounds need to be positively or negatively shifted respectively. Similarly, if mid-range values are sitting in the positive space then we may need to contract both ends. If only high or low values are sitting in the positive space then we may be able to fix the hyperparameter value all together. If the values are spread out then we can see that the hyperparameter plays a larger role in the objective value than those that are concentrated, or alternatively the contracted hyperparameters have a smaller range or are inappropriately bounded. 

\begin{figure}[htbp]
  \centering
  \begin{subfigure}[b]{0.45\textwidth}
    \includegraphics[width=\textwidth, height=5cm]{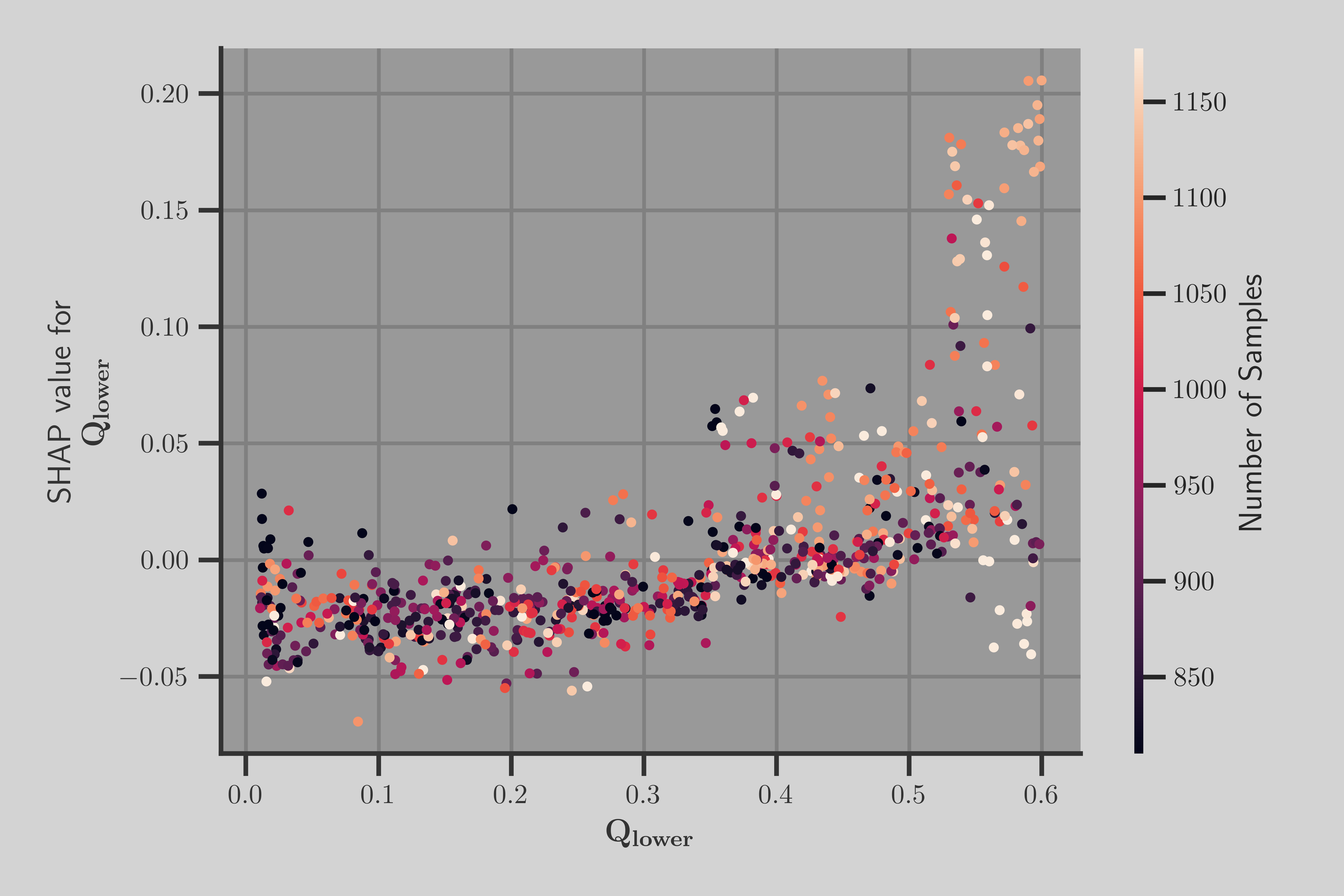}
    \caption{$Q_{lower}$ vs Number of Samples}
    \label{fig:shap:q-lower-samples}
  \end{subfigure}
  \hfill
  \begin{subfigure}[b]{0.45\textwidth}
    \includegraphics[width=\textwidth, height=5cm]{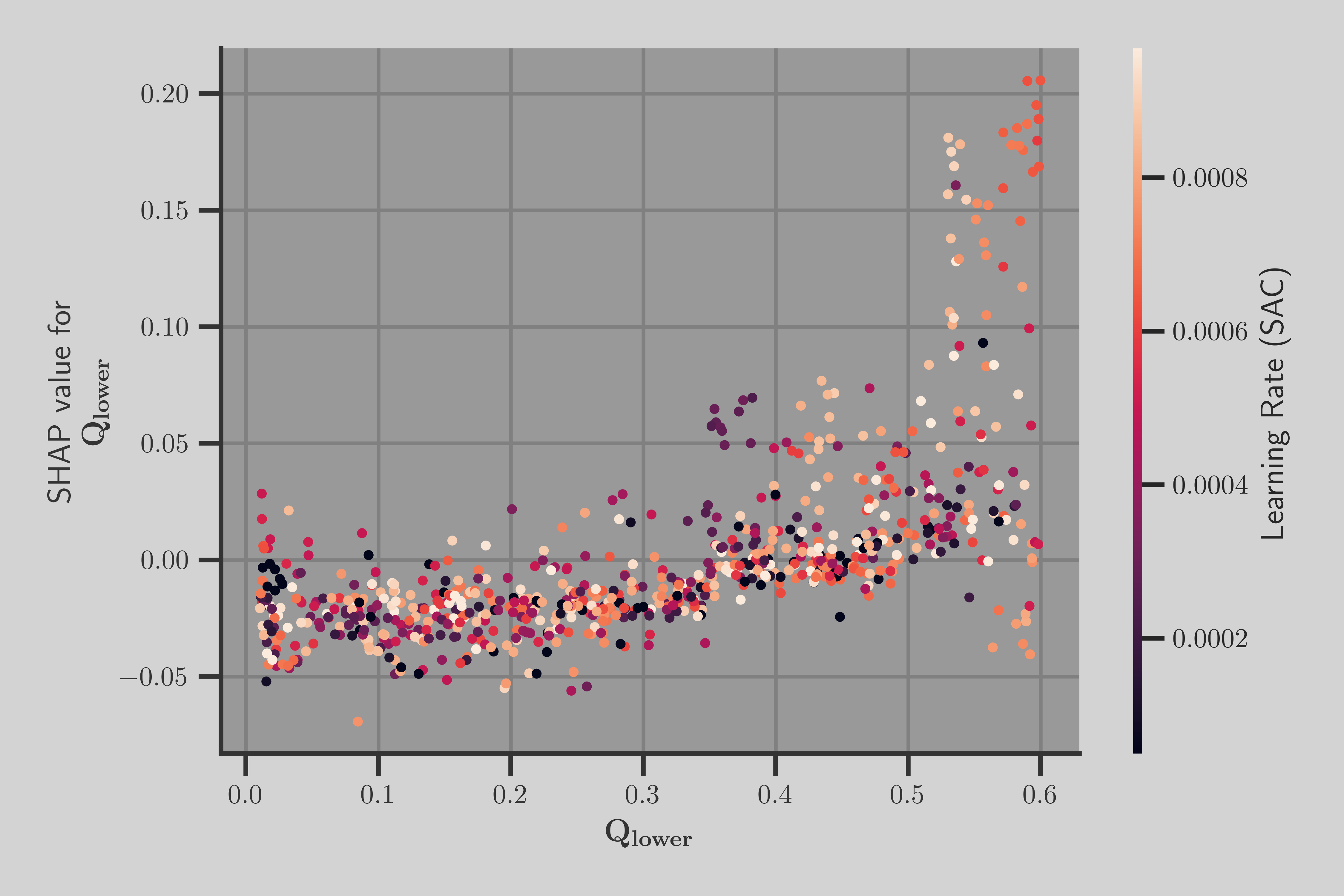}
    \caption{$Q_{lower}$ vs Learning Rate (SAC)}
    \label{fig:shap:q-lower-lr}
  \end{subfigure}

  \begin{subfigure}[b]{0.45\textwidth}
    \includegraphics[width=\textwidth, height=5cm]{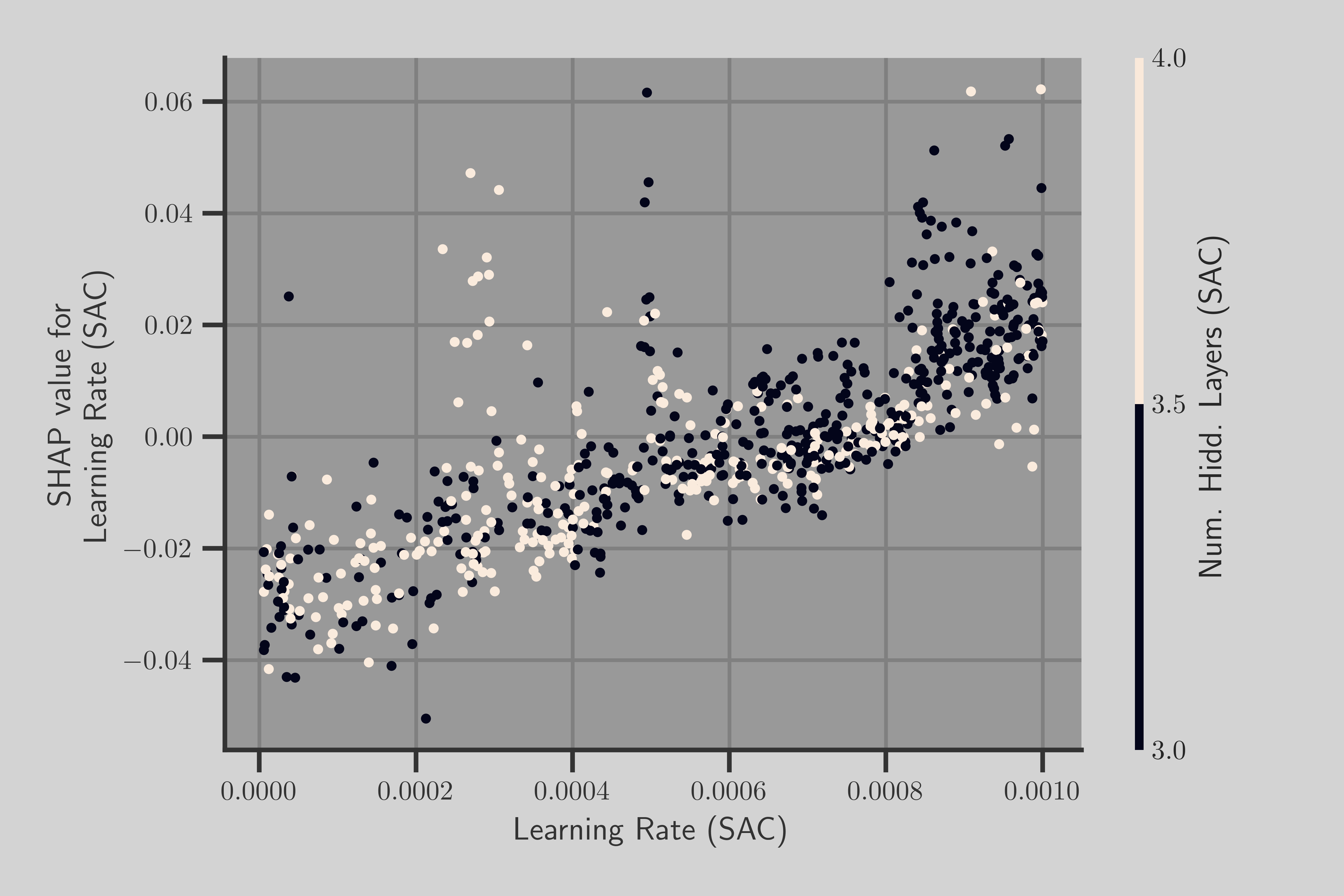}
    \caption{Learning Rate {SAC} vs Hidden Layers (SAC)}
    \label{fig:shap:lr-hlsac}
  \end{subfigure}
  \hfill
  \begin{subfigure}[b]{0.45\textwidth}
      \includegraphics[width=\textwidth, height=5cm]{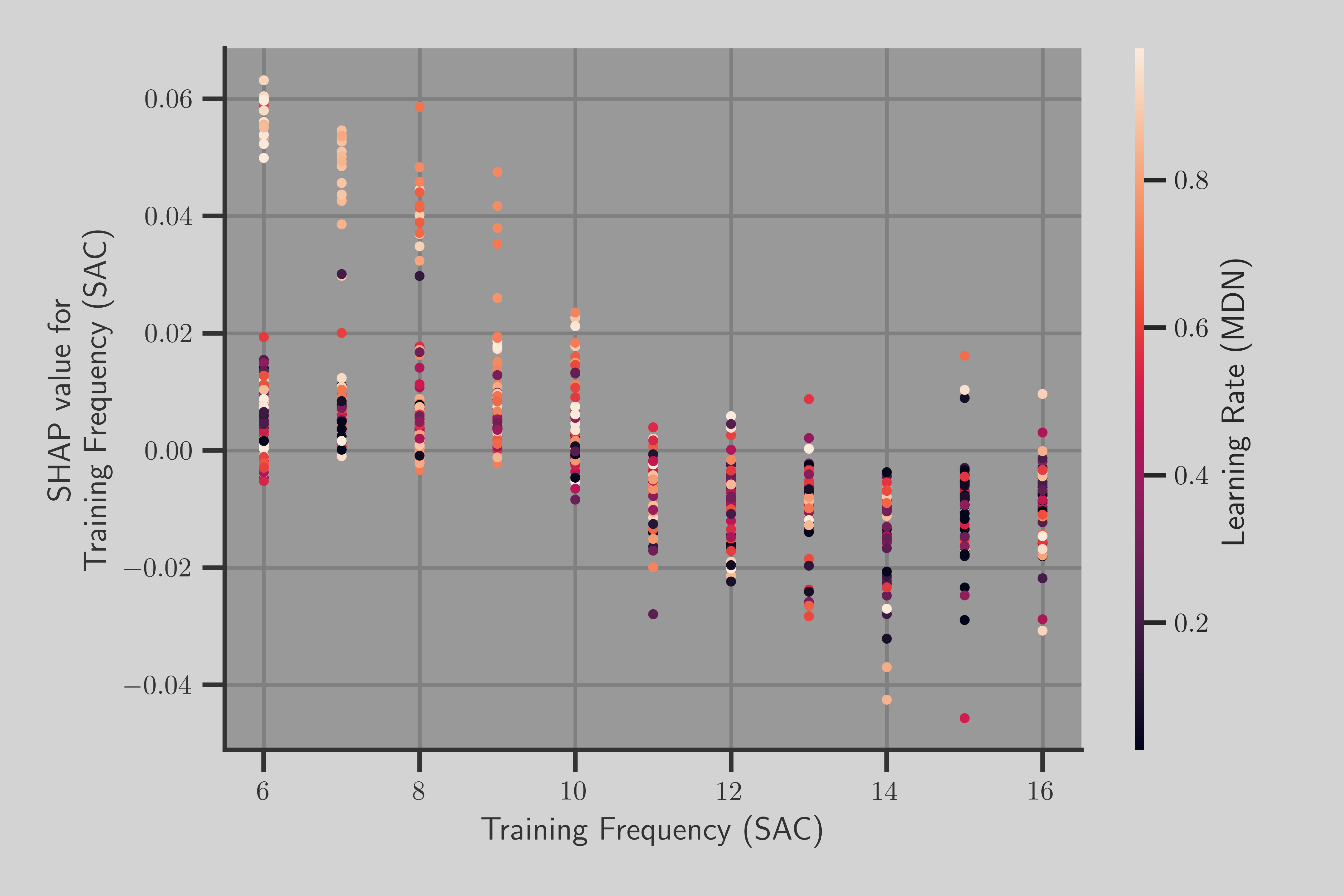}
      \caption{Training Freq. (SAC) vs $Q_{lower}$}
      \label{fig:shap:tfs-q-lower}
  \end{subfigure}
  
  \caption{SHAP Codependency Plots.}
  \label{fig:shap:codep}
\end{figure}

We can see in Figure~\ref{goal:shap:combined_summary} that the $Q_{lower}$ is critical for the performance. With a high spread over the SHAP values with the most positive performers on the high end of the spectrum. This makes sense as the low number of steps during training (150,000) would benefit from sampling less easy and therefore less informative goals. Conversely, $Q_{upper}$ has a lower impact on the objective value, but the best performers are on the lower end of the spectrum. This supports the hypothesis that sampling attainable goals is important as lower $Q_{upper}$ values filters out more difficult goals. Additionally, batch size for the MDN is relatively important with lower values having a large SHAP value. This could be due to the fact that typically, high learning rates were favoured as we can also see in the plot. However, the correlation matrix in Figure~\ref{goal:shap:correlation_matrix} does not show a particularly strong correlation for these values. We can also see that high batch sizes can perform well, but with not as high SHAP values. SAC learning rate shows higher SHAP values for higher values. We could use these SHAP values to inform our hyperparameter bounds for future experiments, e.g. increasing the upper bound for learning rate (SAC) and lowering the MDN batch size.

We can also see that the importance weighting of the parameters in the SHAP plot are broadly consistent with the correlation values in Figure~\ref{goal:shap:correlation_matrix_y}. This is a good sign as it indicates that the SHAP explainer is working as intended. As we are using a random forest regressor with a small number(\~800) data points, we could expect that the model could poorly model the function, so multiple methods of evaluating the importance of hyperparameters is useful. 

Figure~\ref{fig:shap:codep} shows the SHAP codependency plots for the hyperparameters with the strongest cross-correlation values. Figure~\ref{fig:shap:q-lower-samples} shows that the number of samples is positively correlated with $Q_{lower}$, as $Q_{lower}$ increases so too does the SHAP value. Associated with this is a higher number of samples, \~1100. Figure~\ref{fig:shap:q-lower-lr} shows that as $Q_{lower}$ increases the SHAP values increase exponentially, with the highest value performers also having high learning rates for the SAC agent. Figure~\ref{fig:shap:lr-hlsac} and~\ref{fig:shap:tfs-q-lower} have a more obviously linear relationship, higher LR and lower training frequency for the SAC agent correlate to higher SHAP values. This is in conjunction with higher $Q_{lower}$ w.r.t the training frequency of SAC. Three hidden layers with higher learning rates for the SAC agent also correlate to higher SHAP values. 

All the results point to the lower quantile $Q_{lower}$ as being important and performing well with higher values. Indicating that it would be profitable to increase the upper and lower bounds of $Q_{lower}$. Additionally, the learning rate and training frequency of SAC are important for importance, showing that appropriate agent parameters are also important for utilising the curriculum. Another relationship that might be interesting to explore is $Q_{lower}$ and $Q_{upper}$ as these ultimately impact the curriculum's choices and the difficulty of goals for the agent, so it would make sense that they have some codependence. 

\begin{figure}[htbp]
  \centering
  \begin{subfigure}[b]{0.45\textwidth}
    \includegraphics[width=\textwidth, height=5cm]{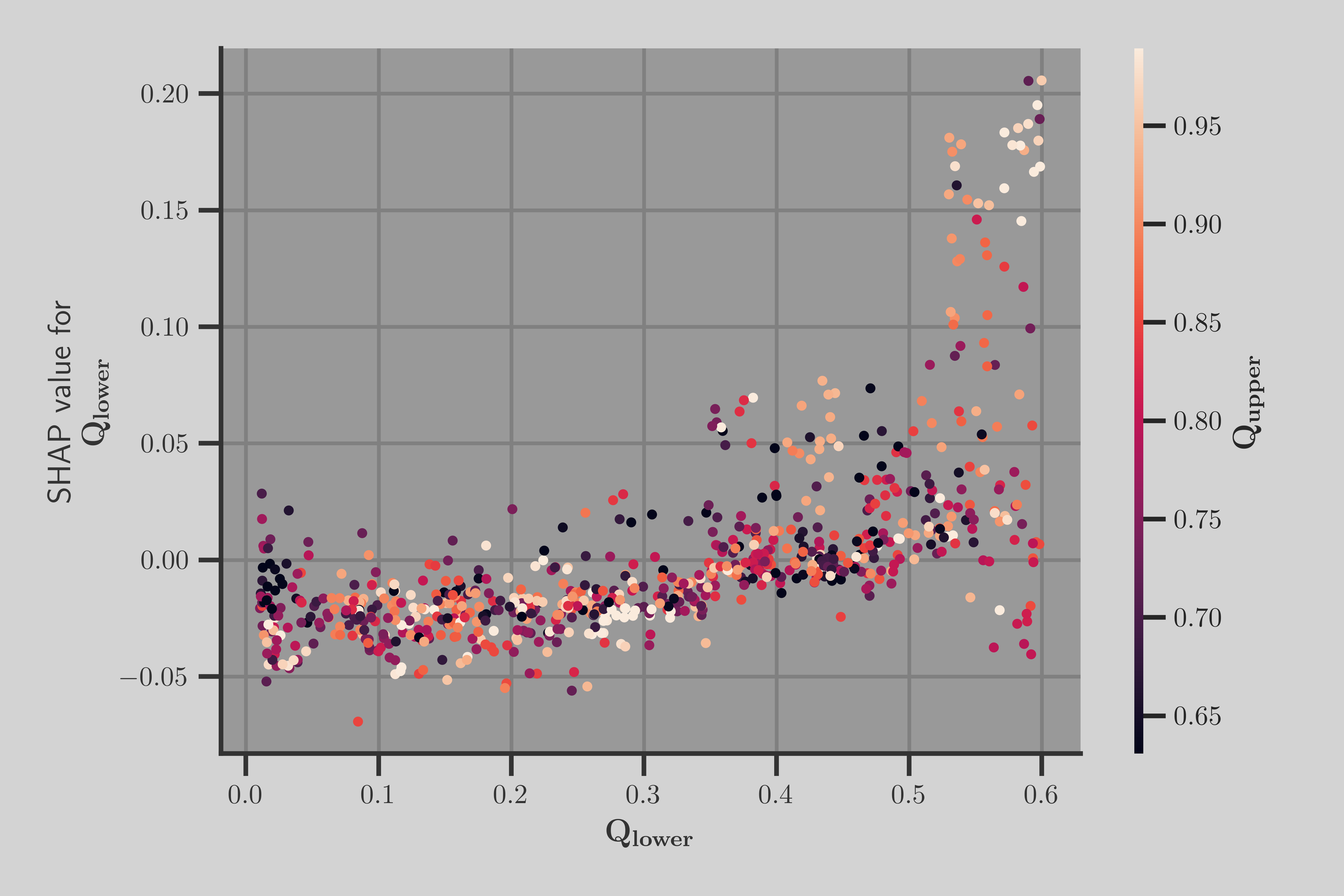}
    \caption{$Q_{lower}$ vs $Q_{upper}$}
    \label{fig:shap:q-lower-q-upper}
  \end{subfigure}
  \hfill
  \begin{subfigure}[b]{0.45\textwidth}
    \includegraphics[width=\textwidth, height=5cm]{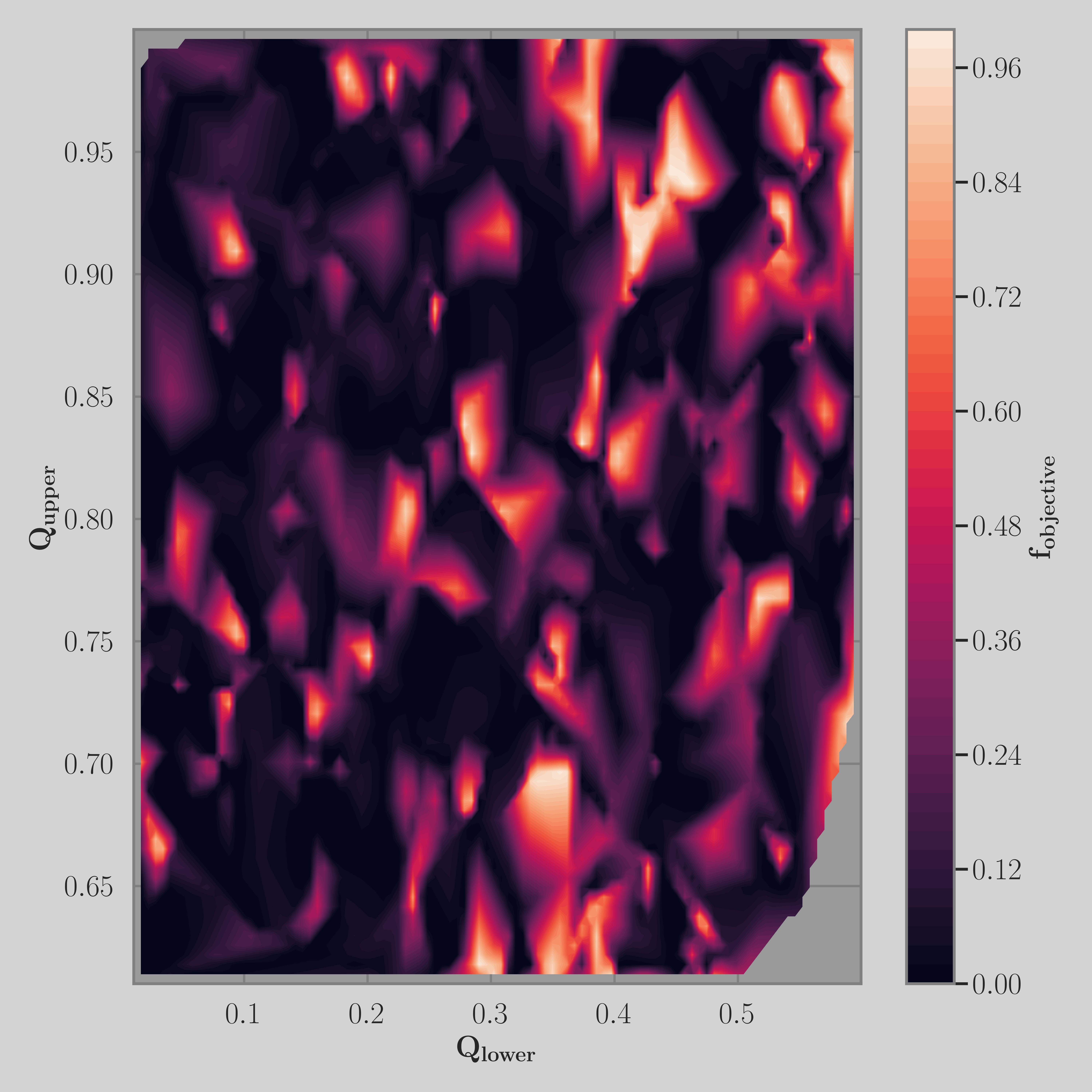}
    \caption{$Q_{lower}$ vs $Q_{upper}$}
    \label{fig:surface:q-lower-q-upper}
  \end{subfigure}  
  \caption{SHAP Codependency and Surface Plots for $Q_{lower}$ vs $Q_{upper}$.}
  \label{fig:qupp:cqlow}
\end{figure}

Figure~\ref{fig:shap:q-lower-q-upper} again shows an exponential increase in SHAP values as $Q_{lower}$ increases. We also see that this is typically coupled with higher $Q_{upper}$ values. This points to the PCL performing better when easier goals are filtered out, but hard ones are maintained. Figure~\ref{fig:surface:q-lower-q-upper} corroborates with this, showing that the most regions of best performers are in the upper right quadrant of the graph.

\section{Conclusion}
In this paper, we have presented a comprehensive analysis of hyperparameter optimisation in reinforcement learning, specifically focusing on the Probabilistic Curriculum Learning (PCL) algorithm. We have demonstrated the importance of hyperparameter tuning in achieving optimal performance and provided practical guidelines for refining hyperparameter search spaces. Our empirical analysis, supported by visualisations and SHAP-based interpretability, highlights the significance of hyperparameter interactions and their impact on RL performance. We present multiple strategies to analyse hyperparameters, their bounds, and their interactions. We show that between our initial and intermediate experiments. 

Our analysis explicitly revealed critical insights, notably highlighting the crucial role of the curriculum hyperparameter $Q_{lower}$ and its interactions with agent-specific hyperparameters like SAC learning rates and training frequency. These results not only confirm the importance of careful hyperparameter tuning in reinforcement learning but also validate the intended function of the PCL curriculum component empirically.

We show that using SHAP by modelling the hyperparameters as a regression problem with respect to the objective value can provide insights into the impacts of hyperparameters and how we could adjust the bounds. To the best of our knowledge, this type of analysis has not been applied to hyperparameter optimisation before and provides a novel approach to understanding the relationships between hyperparameters and their effects on performance.

\newpage
\bibliography{hyperopt}

%%%%%%%%%%%%%%%%%%%%%%%%%%%%%%%%%%%%%%%%%%%%%%%%%%%%%%%%%%%%
\newpage
\appendix

\section{Appendix / supplemental material}
\label{sec:appendix}
\subsection{Hyperparameter Tables}
\begin{table}[htbp]
	\centering
	\begin{tabularx}{\textwidth}{|X|X|X|X|}
		\hline
		\textbf{Name} & \textbf{Lower Bound} & \textbf{Upper Bound} & \textbf{Length} \\
		\hline
		Num Mixtures & 1 & 10 & 1 \\
		\hline
		Hidden Layers & 64 & 512 & [1, 3] \\
		\hline
		Learning Rate & 1e-05 & 0.1 & 1 \\
		\hline
		$\lambda_1$ & 0.8 & 1 & 1 \\
		\hline
		$\lambda_2$ & 0.0 & 0.5 & 1 \\
		\hline
		$\lambda_3$ & 0.0 & 0.5 & 1 \\
		\hline
		Number of samples & 100 & 1000 & 1 \\
		\hline
		$Q_{lower}$ & 0.5 & 0.8 & 1 \\
		\hline
		$Q_{upper}$ & 0.8 & 0.99 & 1 \\
		\hline
    SAC Hidden Layers & 64 & 1024 & [2, 3] \\
		\hline
		SAC Learning Rate & 5e-06 & 5e-5 & 1 \\
        \hline
	\end{tabularx}
	\caption{Initial Hyperparameter Bounds for Experiments.}
	\label{tab:MDNModelparameters}
\end{table}

\begin{table}[htbp]
  \centering
  \begin{tabularx}{\textwidth}{|X|X|X|X|}
		\hline
		\textbf{Name} & \textbf{Lower Bound} & \textbf{Upper Bound} & \textbf{Length} \\
		\hline
		Training Frequency (MDN) & 2 & 10 & 1 \\
		\hline
		Number of Mixtures & 6 & 12 & 1 \\
		\hline
		Layers (MDN) & 64 & 1024 & [3, 4] \\
		\hline
		Learning Rate (MDN) & 0.0001 & 1 & 1 \\
		\hline
		$\mathbf{\lambda_1}$ & 0.85 & 2 & 1 \\
		\hline
		$\mathbf{\lambda_2}$ & 0.1 & 0.5 & 1 \\
		\hline
		$\mathbf{\lambda_3}$ & 0.85 & 2 & 1 \\
		\hline
		$\beta_1$ & 0.0 & 2.0 & 1 \\
		\hline
		$\beta_2$ & 0.0 & 2.0 & 1 \\
		\hline
		$\beta_3$ & 0.0 & 2.0 & 1 \\
		\hline
		Number of Samples & 800 & 1200 & 1 \\
		\hline
		$\mathbf{Q_{lower}}$ & 0.01 & 0.6 & 1 \\
		\hline
		$\mathbf{Q_{upper}}$ & 0.61 & 1 & 1 \\
		\hline
		Batch Size (MDN) & 128 & 1024 & 1 \\
		\hline
		Training Frequency (SAC) & 6 & 16 & 1 \\
		\hline
		Batch Size (SAC) & 700 & 1000 & 1 \\
		\hline
		Layers (SAC) & 100 & 800 & [3, 4] \\
		\hline
		Learning Rate (SAC) & 4e-06 & 0.001 & 1 \\
		\hline
	\end{tabularx}
  \caption{Intermediate Hyperparameter Bounds for Experiments.}
  \label{tab:MDNModel2-parameters}
\end{table}

\begin{table}[H]
	\centering
	\begin{tabularx}{\textwidth}{|X|X|X|X|}
		\hline
		\textbf{Name} & \textbf{Lower Bound} & \textbf{Upper Bound} & \textbf{Length} \\
		\hline
		Training Frequency (MDN) & 1 & 10 & 1 \\
		\hline
		Number of Mixtures & 1 & 12 & 1 \\
		\hline
		Layers (MDN) & 64 & 1024 & [1, 8] \\
		\hline
		Learning Rate (MDN) & 0.0001 & 1 & 1 \\
		\hline
		$\mathbf{\lambda_1}$ & 0.85 & 2 & 1 \\
		\hline
		$\mathbf{\lambda_2}$ & 0.1 & 0.5 & 1 \\
		\hline
		$\mathbf{\lambda_3}$ & 0.85 & 2 & 1 \\
		\hline
		$\beta_1$ & 0.0 & 2.0 & 1 \\
		\hline
		$\beta_2$ & 0.0 & 2.0 & 1 \\
		\hline
		$\beta_3$ & 0.0 & 2.0 & 1 \\
		\hline
		Number of Samples & 900 & 1100 & 1 \\
		\hline
		$\mathbf{Q_{lower}}$ & 0.5 & 0.8 & 1 \\
		\hline
		$\mathbf{Q_{upper}}$ & 0.81 & 1 & 1 \\
		\hline
		Batch Size (MDN) & 128 & 1024 & 1 \\
		\hline
		Training Frequency (SAC) & 1 & 16 & 1 \\
		\hline
		Batch Size (SAC) & 900 & 1000 & 1 \\
		\hline
		Layers (SAC) & 100 & 800 & [1, 8] \\
		\hline
		Learning Rate (SAC) & 1e-04 & 0.1 & 1 \\
		\hline
	\end{tabularx}
	\caption{Final Hyperparameter Bounds for Experiments.}
	\label{tab:SACWeightedMDNExperiment}
\end{table}

\subsection{Correlation Matrix of Hyperparameters}
\begin{figure}[H]
  \centering
  \includegraphics[width=0.75\textwidth, scale=0.75]{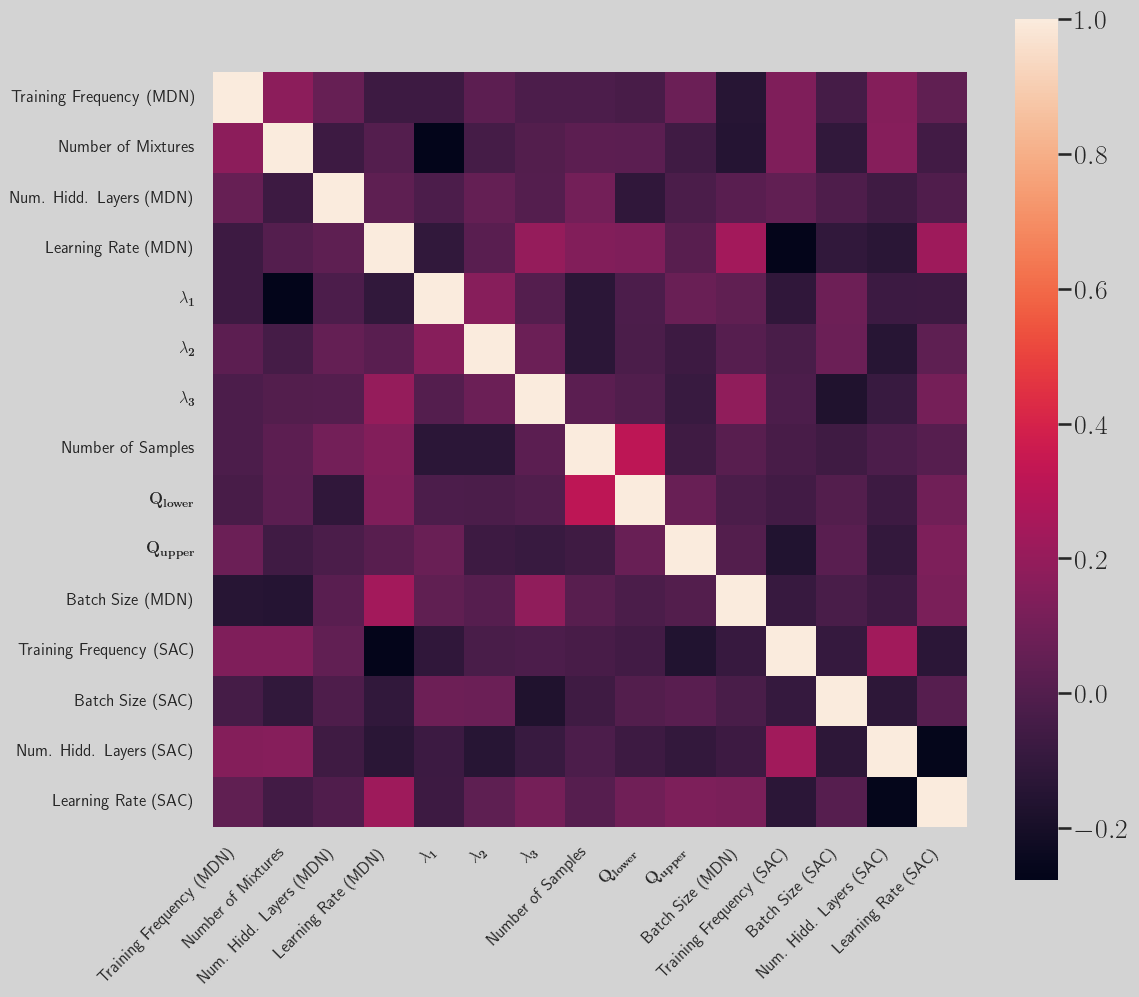}
  \caption{Correlation matrix of hyperparameters for the PCL in the PointMaze environments with no filtering on $f_{objective}$.}
  \label{goal:shap:correlation_matrix}
\end{figure}

\end{document}